\documentclass[journal]{IEEEtran}

\usepackage{cite}
\usepackage{amsmath,amssymb,amsfonts}
\usepackage{graphicx}
\usepackage{placeins}
\usepackage{flafter} 
\usepackage[normalem]{ulem}

\usepackage{dblfloatfix}
\usepackage{array}
\usepackage{tabularx}
\usepackage{multirow}
\usepackage{makecell}
\usepackage{xcolor}
\usepackage{algorithm}
\usepackage{algpseudocode}

\newcommand{\rev}[1]{#1}
\newcommand{\Judor}{JUDOR}

\usepackage{caption}
\algrenewcommand\algorithmicrequire{\textbf{Require:}}
\algrenewcommand\algorithmicensure{\textbf{Ensure:}}
\algrenewcommand\algorithmicfor{\textbf{for}}
\algrenewcommand\algorithmicdo{\textbf{do}}
\algrenewcommand\algorithmicwhile{\textbf{while}}
\algrenewcommand\algorithmicif{\textbf{if}}
\algrenewcommand\algorithmicthen{\textbf{then}}
\algrenewcommand\algorithmicelse{\textbf{else}}
\algrenewcommand\algorithmicend{\textbf{end}}
\algrenewtext{EndFor}{\algorithmicend\ \algorithmicfor}
\algrenewtext{EndWhile}{\algorithmicend\ \algorithmicwhile}
\algrenewtext{EndIf}{\algorithmicend\ \algorithmicif}
\algrenewcommand\algorithmicreturn{\textbf{return}}
\algrenewcommand\algorithmiccomment[1]{\hfill\textit{//}\,#1}

\usepackage[hidelinks,breaklinks,bookmarksnumbered=true,bookmarksopen=true,bookmarksopenlevel=1]{hyperref}
\usepackage{bookmark}

\begin{document}

\title{\texorpdfstring{\rev{JUDOR: CTDE‑Based Multi‑Agent Reinforcement Learning for the Co‑Optimization of Joint UAV Motion and Decentralized Opportunistic Routing in Delay‑Tolerant Networks}}{JUDOR: CTDE-Based Multi-Agent Reinforcement Learning for the Co-Optimization of Joint UAV Motion and Decentralized Opportunistic Routing in Delay-Tolerant Networks}}

\author{Xiao~Wang and Shun-Ren~Yang,~\IEEEmembership{IEEE~Member}%
\thanks{X.~Wang and S.-R.~Yang are with the Department of Computer Science and the Institute of Communications Engineering, National Tsing Hua University, Hsinchu 30013, Taiwan (e-mail: sryang@cs.nthu.edu.tw).}}

\markboth{Wang and Yang: JUDOR for Delay-Tolerant Networks}{Wang and Yang: JUDOR for Delay-Tolerant Networks}

\maketitle

\begin{abstract}
The growing deployment of delay‑tolerant networks (DTNs) has made store‑carry‑forward (SCF) communication indispensable for operating under sparse intermittent connectivity.
\rev{However}, intermittent contact opportunities, finite node buffers, and limited message time‑to‑live (TTL) \rev{degrade} message delivery performance and \rev{induce} network congestion.
\rev{Yet unmanned aerial vehicles (UAVs), with extended radio coverage and controllable trajectories, can bridge ground‑node pairs that would otherwise suffer poor connectivity.}
\rev{Nevertheless}, conventional routing schemes decouple UAV motion control from message replication, and \rev{consequently fail to leverage} UAV mobility to proactively create future contact opportunities.
This paper presents \rev{JUDOR~(\textbf{J}oint‑\textbf{U}AV‑\textbf{D}ecentralized‑\textbf{O}pportunistic‑\textbf{R}outing)}, a centralized training with decentralized execution (CTDE)‑based multi‑agent reinforcement learning framework optimized for joint discrete UAV heading control and decentralized opportunistic routing.
\rev{Specifically}, we model the tightly coupled UAV motion‑routing co‑optimization task as a factored partially observable Markov decision process with a shared team reward.
JUDOR adopts decentralized actors for local‑node execution and a centralized critic used exclusively in training, with optional multi‑horizon congestion‑hotspot auxiliary supervision.
\rev{Through evaluation across four distinct traffic modes}, JUDOR yields consistent performance gains over both classical and learning‑based DTN baselines, including PRoPHET and MaxProp, whilst preserving contact‑limited decentralized execution suitable for real‑world DTN deployments.
\end{abstract}

\begin{IEEEkeywords}
\rev{Centralized Training and Decentralized Execution, Delay-Tolerant Networks, Joint UAV-Routing Optimization, Multi-Agent Reinforcement Learning, Opportunistic Routing, Unmanned Aerial Vehicles}
\end{IEEEkeywords}

\section{Introduction}
\label{sec:intro}
\IEEEPARstart{I}{n} recent years, delay-tolerant networks (DTNs) have attracted extensive research attention as a communication paradigm for environments lacking stable end-to-end connectivity. Typical application scenarios include post-disaster emergency rescue, sparse rural road networks, and remote sensing missions. In these scenarios, terrestrial communication infrastructure is often damaged, overloaded or completely unavailable, leaving vehicles and emergency responders to exchange status and event data without continuous end-to-end communication paths.
Under these conditions, radio links can only be established when nodes move within a limited communication range. Moreover, finite onboard buffer capacity and limited message time-to-live (TTL), together with ground mobility constrained mostly to road networks, further impose physical limits that prevent sustained connectivity in practical deployments.
DTNs cope with these limits by store–carry–forward (SCF): a relay stores a message, carries it while moving, and forwards it when another node comes into range~\cite{badawi2025routing}.
Nevertheless, DTNs cannot dictate the timing of contact opportunities, as node mobility follows pre‑defined trajectories outside protocol control. Pairwise contacts therefore arise purely by chance. Conventional routing protocols are thus restricted to neighbor selection for message replication, with no capacity to proactively generate new encounters.

With the rapid development of unmanned aerial vehicles (UAVs), controllable aerial platforms naturally serve as a complementary solution. UAVs provide extended radio coverage and agile trajectories to bridge disconnected ground clusters that are otherwise limited by road-bound mobility~\cite{du2021routing,fan2025novel}.
UAV assistance, however, still suffers from inherent network‑level limitations. 
Heterogeneous communication ranges between aerial relays and ground nodes mean that even a modest‑size UAV fleet is unable to deliver full network connectivity and can only establish intermittent, range‑limited contacts.
Classical DTN routing protocols, with PRoPHET~\cite{lindgren2004probabilistic} as a representative example, operate under the assumption that \rev{the mobility of a node is controlled by the node itself}. Though applicable to general opportunistic‑forwarding scenarios, such protocols derive delivery probabilities from past encounter records and select relay candidates while treating UAV trajectories as unmodifiable external inputs.
Consequently, by treating UAV mobility and message forwarding as separate tasks, these protocols cannot jointly optimize aerial motion with SCF replication and fail to fully exploit UAV‑generated contact opportunities under highly intermittent network conditions, which motivates joint optimization between UAV motion and SCF‑based message forwarding.

Given such communication‑range limitations, every UAV heading decision reshapes the contact graph for subsequent timesteps, tightly coupling UAV mobility control and packet forwarding. Accordingly, UAV‑assisted DTN control must account for future topology evolution, rather than merely relying on the current encounter set.
\rev{UAV-assisted DTN control} faces three coupled difficulties.
First, \textbf{time‑varying topology}: UAV motion and ground mobility jointly change the contact graph. Routing decisions alter node buffer occupancy and the resulting congestion states, which further guide subsequent UAV heading selections and thus alter future node contacts.
Second, \textbf{team learning and decentralized execution}: cooperative credit assignment (the problem of attributing team‑level collective reward outcomes to the local actions of individual agents) becomes challenging because online agents observe only local and contact‑exchangeable features and lack a global network view~\cite{lowe2017multi,foerster2018counterfactual}.
Third, \textbf{weak and delayed learning signals}: delivery events are sparse relative to forwarding attempts, while congestion hotspots evolve with buffer pressure and message injection, so delivery-centric feedback alone provides limited cues for system-level decisions~\cite{sammou2026pf,du2021routing,wen2024enhancing}.

Existing studies address these difficulties only in isolation.
Classical and heuristic UAV-assisted routers improve encounter-driven relay selection or UAV trajectory planning, yet they do not co-optimize discrete UAV headings with per-node SCF replication~\cite{lindgren2004probabilistic,du2021routing,fan2025novel}.
\rev{Reinforcement learning (RL)--based DTN routers adapt forwarding policies under intermittent connectivity and sparse reward feedback, but remain routing-centric without jointly learning UAV heading control}~\cite{wen2024enhancing,sammou2026pf}.
Cooperative multi-agent schedulers coordinate routing or relay decisions on contact graphs and multi-UAV platforms, yet typically omit a complete SCF model of buffer-limited replication and delivery-centric team rewards~\cite{zhou2025distributed,fan2024delay}.
A unified treatment that jointly resolves time-varying topology coupling, contact-limited decentralized execution with team credit assignment, \rev{as well as weak and delayed learning signals} is therefore still lacking.

These \rev{unsolved} gaps motivate \textbf{JUDOR}~(\textbf{J}oint-\textbf{U}AV-\textbf{D}ecentralized-\textbf{O}pportunistic-\textbf{R}outing), a discrete-time SCF multi-agent cooperative-control system under factored \rev{centralized training with decentralized execution (CTDE)} policies.
\rev{Its core design goal is to co‑learn discrete UAV headings and per‑node SCF replication under one cooperative team objective.}
Our contributions are as follows, each aligned with one of the three coupled difficulties above.
\begin{itemize}
\item \textbf{Episodic optimization and joint decision‑process formulation:}
To address the first challenge of time‑varying topology, we formalize joint UAV–routing control as a cooperative episodic optimization problem. This problem is subject to SCF constraints and accounts for delivery, congestion, and UAV mobility metrics. We then model it as a factored partially observable Markov decision process (POMDP) that captures sequential routing–motion coupling. We further derive a per‑step reward proxy under contact‑limited observations, enabling joint treatment of mutually dependent UAV motion evolution and message replication.
\item \textbf{Policy factorization with decentralized online execution:} To address the second challenge of decentralized execution and team learning, we factor \Judor{} into decentralized routing and UAV heading actors that act only on local on‑board and contact‑exchangeable inputs, together with a centralized critic that leverages global training‑time statistics to attribute collective team returns to individual agent actions. This enables effective multi‑agent coordination without requiring global network states during online node execution.
\item \textbf{Modular auxiliaries for sparse delayed feedback:} To address the third challenge of weak and delayed learning signals, we equip \Judor{} with optional congestion-aware auxiliary mechanisms, including multi-horizon congestion-hotspot prediction and heading-congestion-hotspot alignment rewards, and systematically evaluate their benefits and drawbacks across diverse message-injection modes.
\end{itemize}
This paper is organized as follows. Sec.~\ref{sec:related} presents multi-agent RL preliminaries (Sec.~\ref{sec:related:prelim}) and related work (Sec.~\ref{sec:related:works}). Sec.~\ref{sec:system} describes the system model. Sec.~\ref{sec:problem} formulates the episodic objective and factored POMDP, and Sec.~\ref{sec:method} develops the proposed \Judor{} framework. Sec.~\ref{sec:experiments} reports simulation results. Finally, Sec.~\ref{sec:conclusion} draws conclusions for this work.
\section{Preliminaries and Related Work}
\label{sec:related}
\subsection{Preliminaries: Multi‑Agent RL Background}
\label{sec:related:prelim}
\label{sec:prelim}
We first summarize the multi‑agent reinforcement learning concepts that underpin the subsequent formulation and algorithm design.

\textbf{Factored partially observable MDP.}
Let $T_{\max}$ denote the number of time slots in one episode, and let the time index be $t\in\{0,1,\ldots,T_{\max}\}$. A cooperative multi‑agent task over this finite horizon can be cast as a partially observable Markov decision process (POMDP): all agents optimize a shared team reward, yet each agent observes only a local portion of the underlying state.
Let $\mathbf{a}^{(t)}\triangleq(a_1^{(t)},\ldots,a_M^{(t)})$ and $\mathbf{o}^{(t)}\triangleq(o_1^{(t)},\ldots,o_M^{(t)})$ denote the joint action and joint observation over $M$ decision units at step $t$.
When control is factored, the joint policy maps $\mathbf{o}^{(t)}$ to a distribution over $\mathbf{a}^{(t)}$ and factorizes into a product of per-unit conditional distributions, each selecting $a_n^{(t)}$ from its local observation $o_n^{(t)}$ alone.
We write $\pi_\theta$ for the induced joint policy with shared parameter vector $\theta$, and $\pi$ for a policy without a particular parameterization.
The next environment state, however, still depends on the joint action of all units under one common global state that no individual unit fully observes.

\textbf{Centralized training with decentralized execution (CTDE).}
Independent learners treat co‑agents as non‑stationary environment noise, so a shared team reward cannot be reliably attributed to individual actions and training often becomes unstable.
CTDE~\cite{lowe2017multi,foerster2018counterfactual} mitigates this gap by training a centralized critic on global or summarized state statistics while retaining decentralized actors that act from local observations at execution time.
At deployment, only the actors are retained. The critic is used exclusively during training and is not required at inference.
This separation is particularly suitable for DTNs, where intermittent contacts cannot carry global state to every node during inference.

\textbf{Proximal policy optimization (PPO)}~\cite{schulman2017proximal}.
Proximal policy optimization (PPO) is an on‑policy actor–critic reinforcement‑learning algorithm designed to avoid unstable policy updates by constraining the magnitude of policy change. It optimizes the policy via a clipped probability‑ratio surrogate objective, while a critic \(V_\psi\) with parameters \(\psi\) simultaneously fits a state‑dependent value baseline. The critic provides value estimates for generalized advantage estimation (GAE), which computes the advantage $\hat{A}^{(t)}$: a quantity indicating whether the taken action outperforms the expected return of the current state. The probability ratio and clipped surrogate loss are given by
\begin{equation}
\varrho^{(t)}(\theta)
=
\frac{\pi_\theta\!\left(\mathbf{a}^{(t)}\mid\mathbf{o}^{(t)}\right)}
{\pi_{\theta^{\prime}}\!\left(\mathbf{a}^{(t)}\mid\mathbf{o}^{(t)}\right)},
\label{eq:prelim_ppo_ratio}
\end{equation}
\begin{equation}
\begin{aligned}
L_{\mathrm{clip}}(\theta)
=
\mathbb{E}\Bigl[
\min\bigl(
&\varrho^{(t)}(\theta)\hat{A}^{(t)},\\
&\operatorname{clip}\bigl(\varrho^{(t)}(\theta),1-\epsilon,1+\epsilon\bigr)\hat{A}^{(t)}
\bigr)
\Bigr],
\end{aligned}
\label{eq:prelim_ppo_clip}
\end{equation}
where $\pi_{\theta}$ is the target policy to be updated, and $\pi_{\theta'}$ is the fixed behavior policy that generates trajectories for the on‑policy training batch. $\epsilon>0$ denotes the clip radius limiting the allowable divergence between $\pi_\theta$ and $\pi_{\theta'}$. Though originally proposed for single‑agent settings, this clipped surrogate objective can be reused to optimize a factored joint multi‑agent policy under the CTDE paradigm. All agents maximize the same shared team‑level reward.

\subsection{Related Work}
\label{sec:related:works}
This section reviews four major research threads relevant to our problem domain, prior to positioning the proposed \Judor{} framework against existing literature.

\subsubsection{Classical DTN Routing}
\label{sec:related:dtn}
This subsection surveys classical delay‑tolerant network routing based on heuristic forwarding schemes.
DTN routing protocols span flooding and epidemic families, quota‑based spray methods, and heuristic‑scoring forwarders~\cite{spyropoulos2010routing}.
Heuristic‑scoring schemes such as PRoPHET~\cite{lindgren2004probabilistic} and MaxProp~\cite{burgess2006maxprop} rank next hops from encounter histories, meeting recency, and buffer pressure. Recent classical variants add cache‑state awareness~\cite{zhang2025epidemic} or social/geographic context~\cite{fernandes2025social,ullah2025sr}.
Comparative studies further contrast statistical, analytical, and learning‑based DTN routers under contact‑plan uncertainty~\cite{d2025comparing}.
These works optimize forwarding scores while treating relay trajectories as predefined and therefore cannot actively reshape future network topology. This limitation is the first coupled difficulty outlined in the introduction.

\subsubsection{UAV‑Assisted DTN Message Routing}
\label{sec:related:uav}
We next examine existing research on UAV‑aided delay‑tolerant networking, where aerial relays are deployed to enlarge communication opportunities for ground nodes.
UAV relays enlarge contact opportunities through extended range and controllable trajectories, yet message routing must still account for heterogeneous mobility, link persistence, and fleet coordination~\cite{zhou2025review,badawi2025routing}.
Du et al.~\cite{du2021routing} weight meeting probability and persistent connection time in UAV‑assisted vehicular DTNs. Fan et al.~\cite{fan2025novel} rank relays with multi‑attribute utilities and dynamic buffer prioritization for post‑disaster VANETs.
Disaster‑targeted UAV–DTN route discovery~\cite{chakrabarti2025noble} further improves path construction under aerial assistance.
These protocols refine encounter scoring, trajectory utilities, or path construction, but they decouple flight planning from decentralized per‑node SCF decisions. As a result, they still fall short of fully addressing the first coupled difficulty: they do not jointly optimize UAV headings together with per‑node message replication under road‑constrained mobility and heterogeneous communication ranges.

\subsubsection{Learning‑Based DTN Routing}
\label{sec:related:learning}
Learning‑based DTN routing extends classical opportunistic forwarding with tabular and deep reinforcement learning~\cite{mammeri2019reinforcement,salam2024constructive}.
Representative designs such as FQLRP~\cite{dhurandher2020reinforcement}, Q‑learning spray‑and‑wait variants~\cite{yao2024improved}, and disaster‑recovery DRL routers~\cite{wen2024enhancing} adapt forwarding decisions using buffer status, mobility traces, or social‑context proxies.
Q‑learning anycast routing over multi‑base‑station UAV graphs~\cite{xiang2025performance} likewise learns relay selection under aerial mobility, while predictive routers such as PF‑DTN~\cite{sammou2026pf} couple LSTM trajectory forecasting with hybrid relay ranking.
Despite these adaptive gains, the dominant pattern remains \emph{routing‑only} control: per‑node forwarding may adapt, but UAV headings are not co‑optimized jointly with replication inside a shared store‑carry‑forward (SCF) simulation environment, leaving the joint mobility‑routing coupling unaddressed.

\subsubsection{Cooperative multi‑agent reinforcement learning (MARL) for Networking and Aerial Relays}
\label{sec:related:marl}
We now turn to cooperative multi‑agent reinforcement learning (MARL) techniques applied to network control and aerial relay systems.
Recent GNN‑based multi‑agent DRL for interplanetary networks~\cite{zhou2025distributed,tian2025scaling} applies PPO with graph attention for distributed routing and scheduling on planned contact graphs.
Multi‑UAV MADRL for wireless relay networks~\cite{fan2024delay} co‑optimizes trajectory and transmission but models forwarding as generic relay service rather than decentralized message replication with buffer pressure and delivery‑centric team rewards.
These lines scale cooperative control on graphs or aerial relays, yet they rarely jointly‑learn discrete UAV headings alongside per‑node SCF actions under real DTN operating conditions.
Auxiliary loss functions are known to stabilize representation learning when task rewards are sparse~\cite{jaderberg2016reinforcement}. Nevertheless, existing MARL‑aided networking works seldom tackle the second difficulty of team‑credit assignment under purely decentralized online execution, nor do they explicitly cope with the third challenge arising from sparse and delayed delivery feedback in SCF systems.

\begin{table*}[!t]
\caption{Positioning of \Judor{} relative to four representative research lines.
The columns record joint UAV--routing control,  SCF simulation, CTDE or MARL, a congestion-hotspot alignment, and contact-limited local deployment.
\checkmark: core native capability.
(\checkmark): partial or conditionally effective.
--: not supported.}
\label{tab:related_positioning}
\centering
\scriptsize
\setlength{\tabcolsep}{2.5pt}
\renewcommand{\arraystretch}{1.12}
\begin{tabular*}{\textwidth}{@{\extracolsep{\fill}}lccccc@{}}
\hline
Research line & Joint UAV and routing & SCF simulation & CTDE/MARL & Congestion-hotspot auxiliary & Local deployment \\
\hline
Classical DTN routing~\cite{lindgren2004probabilistic,burgess2006maxprop} & -- & (\checkmark) & -- & -- & \checkmark \\
Heuristic UAV DTN/VANET~\cite{du2021routing,fan2025novel,chakrabarti2025noble} & (\checkmark) & (\checkmark) & -- & -- & \checkmark \\
Learning‑based DTN routing~\cite{dhurandher2020reinforcement,yao2024improved,wen2024enhancing,xiang2025performance,sammou2026pf} & -- & (\checkmark) & (\checkmark) & (\checkmark) & (\checkmark) \\
Generic MARL routing~\cite{zhou2025distributed,fan2024delay} & (\checkmark) & -- & \checkmark & -- & (\checkmark) \\
\textbf{\Judor{} (this work)} & \checkmark & \checkmark & \checkmark & \checkmark & \checkmark \\
\hline
\end{tabular*}
\end{table*}

\subsubsection{Paper Positioning Against Existing Works}
\label{sec:related:positioning}
None of the four research threads surveyed above simultaneously resolves all three coupled difficulties raised in the introduction. Table~\ref{tab:related_positioning} summarizes their capability coverage.
Built upon a unified discrete‑time SCF system under the CTDE‑PPO paradigm (Sec.~\ref{sec:related:prelim}), \Judor{} jointly optimizes per‑node forwarding and UAV headings while supporting contact‑limited decentralized inference for practical deployment.

\section{System Model}
\label{sec:system}

First, we present the network architecture and physical models. Second, we define the joint routing and UAV motion control model. Third, we characterize UAV fleet states and congestion metrics for UAV motion control.

\subsection{Network Architecture and Physical Models}
\label{sec:system:model}
\begin{figure*}[!t]
\centering
\begin{minipage}[t]{0.49\textwidth}
\centering
\includegraphics[width=\linewidth]{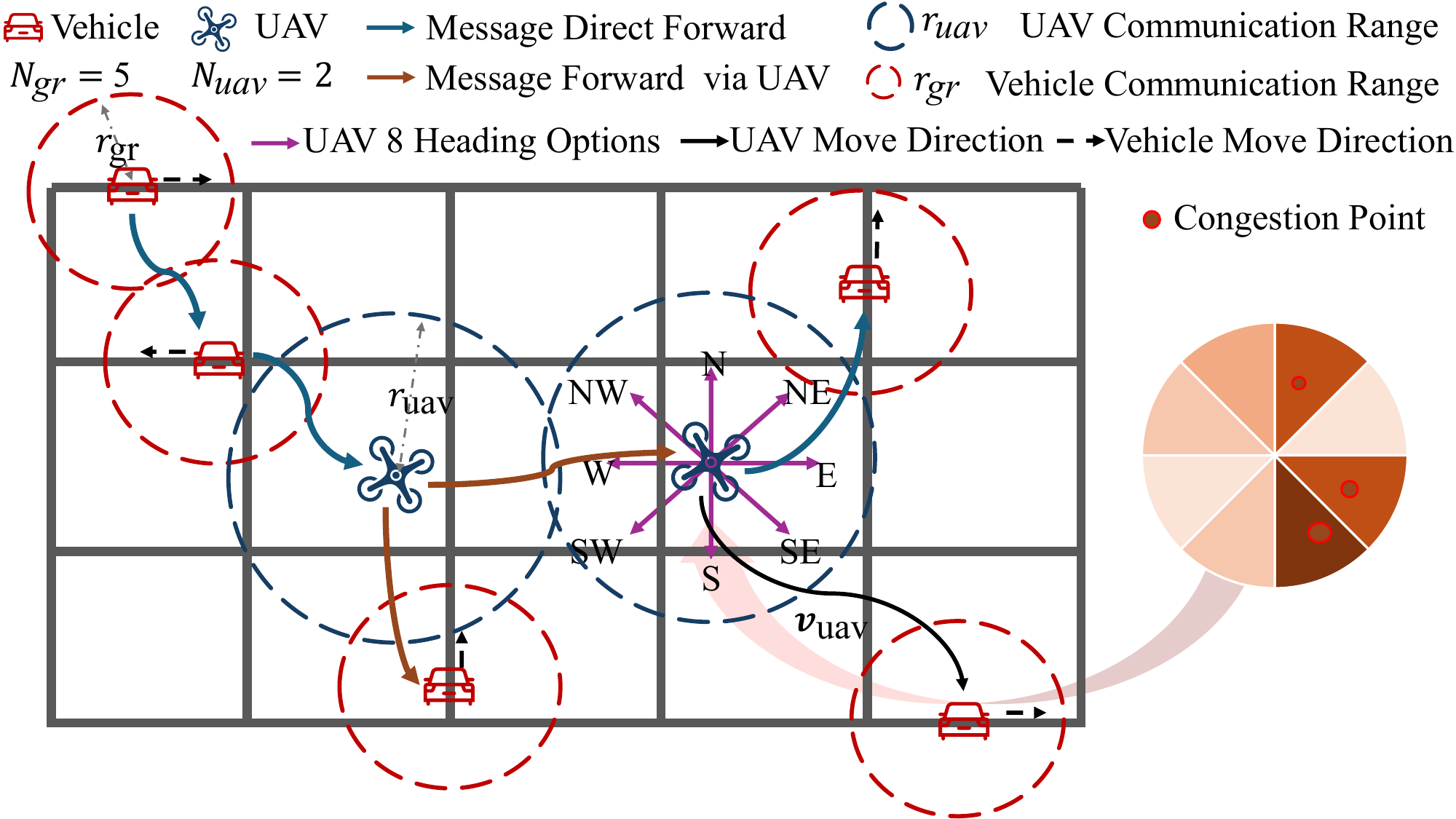}
\end{minipage}\hfill
\begin{minipage}[t]{0.49\textwidth}
\centering
\includegraphics[width=\linewidth]{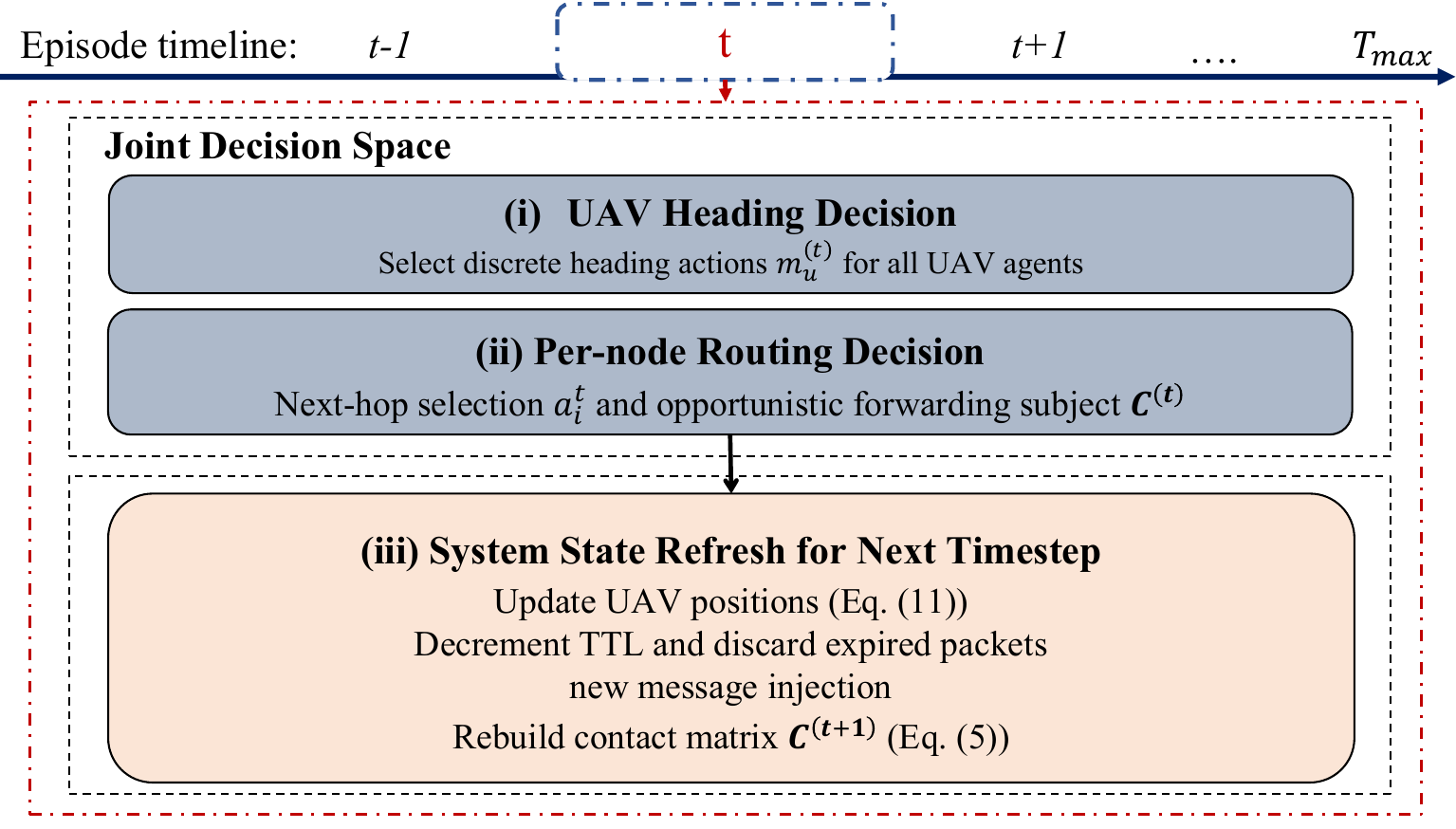}
\end{minipage}
\caption{System model: (a)~spatial layout of ground and UAV nodes; (b)~per-step execution pipeline.}
\label{fig:system_model}
\end{figure*}

In Fig.~\ref{fig:system_model}, we illustrate the SCF DTN architecture, which combines road‑constrained ground vehicles and controllable UAV relays operating on a 2‑D plane. Under this architecture, all nodes perform opportunistic message replication subject to range‑limited communication contacts and finite per‑node buffer capacity.

Each node maintains a local message queue subject to a maximum buffer capacity $B_b$, and the UAV relay fleet furnishes aerial contact opportunities whose spatial distribution is shaped by discrete heading decisions.
We consider the network to consist of $N{=}N_{\mathrm{gr}}{+}N_{\mathrm{uav}}$ nodes indexed by $i\in\{0,1,\ldots,N{-}1\}$, where ground vehicles occupy indices $\{0,\ldots,N_{\mathrm{gr}}{-}1\}$ and the $u$-th UAV relay is mapped to index $i_u{=}N_{\mathrm{gr}}{+}u$.
Communication feasibility at step $t$ is encoded by the contact matrix $\mathbf{C}^{(t)}=\bigl[C_{ij}^{(t)}\bigr]$, whose $(i,j)$-th entry $C_{ij}^{(t)}\in\{0,1\}$ indicates whether nodes $i$ and $j$ are in contact, and message replication is permitted only when $C_{ij}^{(t)}{=}1$.
We adopt the discrete‑time slotted structure of Sec.~\ref{sec:prelim}: each episode runs over the $T_{\max}$ time slots indexed by $t\in\{0,1,\ldots,T_{\max}\}$.
Fig.~\ref{fig:system_model}(a) illustrates the spatial domain at time step $t$. Within this domain, road-constrained ground vehicles travel at speed $v_{\mathrm{gr}}$ with communication range $r_{\mathrm{gr}}$, while freely moving UAV relays fly at speed $v_{\mathrm{uav}}$ with communication range $r_{\mathrm{uav}}$.
In our scenario, ground mobility is modeled as a predefined process independent of our algorithm control, whereas UAV headings and opportunistic forwarding constitute the joint controllable decision space optimized by our system.

\textbf{1) Node Mobility Model:}
We assume all nodes are equipped with onboard Global Navigation Satellite System (GNSS) localization. Ground vehicles are confined to a predefined road network, and each travels at the same constant speed \(v_{\mathrm{gr}}\).
For aerial relay nodes, the mobility model differs substantially. UAV relays move freely on the two-dimensional plane at constant speed $v_{\mathrm{uav}}$. At each time step, each relay selects a discrete heading and travels a distance $v_{\mathrm{uav}}$ along the chosen direction.
Consequently, ground vehicles provide routing decisions only, whereas each UAV simultaneously participates in SCF replication and controllable aerial motion.

\textbf{2) Communication Contact Model:}
\label{sec:system:contacts}
Let $\mathbf{p}_i^{(t)}=(p_{i,x}^{(t)},p_{i,y}^{(t)})\in\mathbb{R}^2$ denote the planar coordinate of node $i$ at time step $t$, and let $\mathbf{v}_i^{(t)}\in\mathbb{R}^2$ denote its planar velocity.
The Euclidean distance between nodes $i$ and $j$ is defined as
\begin{equation}
d_{ij}^{(t)} = \bigl\| \mathbf{p}_i^{(t)} - \mathbf{p}_j^{(t)} \bigr\|_2.
\label{eq:distance}
\end{equation}
We employ a range-limited contact model in which the effective range $r_{ij}$ depends on the node types of the two endpoints:
\begin{equation}
r_{ij}=
\begin{cases}
r_{\mathrm{gr}}, & \text{both } i \text{ and } j \text{ are ground vehicles};\\[2pt]
r_{\mathrm{uav}}, & \text{at least one endpoint is a UAV}.
\end{cases}
\label{eq:comm_range}
\end{equation}
At time step $t$, a communication link between nodes $i$ and $j$ is established if and only if $d_{ij}^{(t)}\le r_{ij}$.
Instantaneous pairwise connectivity is characterized by binary contact indicators $C_{ij}^{(t)}\in\{0,1\}$, which constitute the time-varying contact matrix $\mathbf{C}^{(t)}=\bigl[C_{ij}^{(t)}\bigr]_{i,j=0}^{N-1}$:
\begin{equation}
C_{ij}^{(t)}=
\begin{cases}
1, & i\neq j \text{ and } d_{ij}^{(t)}\le r_{ij};\\[2pt]
0, & \text{otherwise}.
\end{cases}
\label{eq:contact_matrix}
\end{equation}
A diagonal constraint $C_{ii}^{(t)}{=}0$ is imposed for all $i$ to eliminate self-loop contacts.
The raw contact degree of node $i$ counts the number of reachable neighbors at step $t$,
\begin{equation}
\nu_i^{(t)} = \sum_{j=0}^{N-1} C_{ij}^{(t)},
\label{eq:contact_degree}
\end{equation}
and the normalized contact degree
\begin{equation}
\bar{\nu}_i^{(t)} = \frac{\nu_i^{(t)}}{N-1} \in [0,1],
\label{eq:norm_degree}
\end{equation}
which measures neighbor density within the unit interval.

\textbf{3) Message Traffic, Buffering, and Forwarding Model:}
\label{sec:system:messages}
Messages are injected by a traffic generation process that is independent of per-step routing and UAV actions. Each message records its source, destination, payload size $s_q$ with maximum admissible size $S_{\max}$, hop count, and remaining TTL.

Due to the limited buffer space at each node, unrestrained replication may induce buffer saturation and delivery failure. Therefore, optimal forwarding must incorporate real-time buffer occupancy during packet replication operations.
The buffer utilization of node $i$ at time step $t$ is defined as
\begin{equation}
b_i^{(t)} = \frac{\bigl|\mathcal{B}_i^{(t)}\bigr|}{B_b} \in [0,1],
\label{eq:buf_util}
\end{equation}
where $\mathcal{B}_i^{(t)}$ denotes the set of messages buffered at node $i$.

All forwarding and buffer updates comply with the following constraints:
\begin{enumerate}
    \item \textbf{Contact Feasibility}: A message transfer from node $i$ to node $j$ at time step $t$ is permitted only if $C_{ij}^{(t)}{=}1$.
    \item \textbf{Single Reception Constraint}: Each node accepts at most one replicated message per time step.
    \item \textbf{Buffer Admission Rule}: A replicated message is accepted if the receiver has available buffer space. When the buffer is full, the oldest stored message is dropped prior to admitting the incoming replicated message.
    \item \textbf{Replication and Delivery Logic}: A valid routing action replicates the selected message to an eligible neighboring node. Delivery is recorded once the destination is reached, and the hop counter is incremented on each forwarding operation.
    \item \textbf{TTL Decay Mechanism}: After all forwarding operations complete at time step $t$, the TTL of every buffered message is decremented and expired messages are purged.
\end{enumerate}

\subsection{Joint Routing and UAV Motion Control Model}
\label{sec:system:control}

Building on the mobility, contact, and message models above, we define joint per-node forwarding and discrete UAV heading control as the integrated action space, enabling the policy to adjust future contact opportunities.
Under intermittent connectivity, each node must select at most one replication action from a candidate set constructed based on its local buffer and instantaneous contacts, denoted by $\{j:C_{ij}^{(t)}{=}1\}$, while each UAV relay additionally selects a heading bin that updates its planar position and thereby modifies $\mathbf{C}^{(t+1)}$.

Let $K$ denote the maximum number of eligible forwarding candidates maintained per node.
Each candidate is a \emph{(message, next-hop neighbor)} pair ranked according to three priority criteria:
(a) one-hop direct delivery when the neighbor is identical to the message destination;
(b) shorter Euclidean distance from the neighbor to the target destination for non-direct transfers;
(c) smaller remaining TTL as tiebreaker.
Only the top-$K$ ranked candidates are retained after truncation.

Each node $i\in\{0,\ldots,N{-}1\}$ selects a routing action from the discrete action space
\begin{equation}
a_i^{(t)} \in \mathcal{A}_R = \{0,1,\ldots,K\},
\label{eq:routing_action}
\end{equation}
where index $0$ denotes the idle operation and indices $1$ to $K$ correspond to prioritized forwarding candidates.
Ground vehicle positions evolve along predefined road trajectories without controllable motion inputs.

UAV relays  $u\in\{0,\ldots,N_{\mathrm{uav}}{-}1\}$ are appended after the $N_{\mathrm{gr}}$ ground nodes.
In addition to routing decision $a_{i_u}^{(t)}\in\mathcal{A}_R$, each relay selects a discrete heading action
\begin{equation}
m_u^{(t)} \in \mathcal{A}_M = \{0,1,\ldots,B-1\},
\label{eq:heading_action}
\end{equation}
with fixed bin number $B{=}8$, where each bin $k\in\{0,\ldots,B{-}1\}$ is associated with a time-invariant unit direction \(\mathbf{d}_k\) along one of the eight compass orientations.
The set \(\{\mathbf{d}_k\}_{k=0}^{B-1}\) is a fixed dictionary of the heading action space and does not depend on \(t\).
Selecting action \(m_u^{(t)}\) therefore amounts to choosing one entry of this dictionary.
The joint control tuple $(a_{i_u}^{(t)},\,m_u^{(t)})\in\mathcal{A}_R\times\mathcal{A}_M$ is generated independently at each step, and the UAV position update follows
\begin{equation}
\mathbf{p}_{i_u}^{(t+1)} = \mathbf{p}_{i_u}^{(t)} + v_{\mathrm{uav}} \,\mathbf{d}_{m_u^{(t)}}.
\label{eq:uav_motion}
\end{equation}

By combining per-node message replication with UAV heading assignment, we derive the step-wise execution pipeline visualized in Fig.~\ref{fig:system_model}(b), which follows three sequential stages:
(a) Execute all UAV heading actions \(m_u^{(t)}\);
(b) Carry out per-node routing actions \(a_i^{(t)}\) under the contact matrix \(\mathbf{C}^{(t)}\) and aforementioned buffer constraints;
(c) Refresh UAV positions according to Eq.~\eqref{eq:uav_motion}, decrement message TTL values, generate packet arrivals where applicable, and recalculate the contact matrix \(\mathbf{C}^{(t+1)}\) alongside the set of eligible forwarding candidates.

\subsection{UAV State Characterization and Congestion Metrics}
\label{sec:system:stress}

We characterize UAV fleet state and define congestion metrics for UAV heading control. We first define fleet dispersion $\xi$ and heading-switch ratio $\kappa$, then quantify scalar delivery stress on ground nodes, next construct in-range congestion density and sector-wise congestion, and finally define UAV navigation observation features (Sec.~\ref{sec:system:uav_nav}), culminating in the navigation vector $\mathbf{z}_u^{(t)}$ consumed by the UAV heading policy.

\textbf{1) UAV Fleet Dispersion and Heading-Switch Ratio:}
\label{sec:system:uav_dispersion}
\label{sec:system:uav_geometry}
Under decentralized heading control, independent UAV motion decisions may induce relay clustering or heading action oscillations (frequent bin switches).
We define the pairwise normalized separation between distinct relays $u$ and $v$ as
\[
\zeta_{uv}^{(t)} = \min\left(1,\, \frac{\left\| \mathbf{p}_{i_u}^{(t)} - \mathbf{p}_{i_v}^{(t)} \right\|_2}{r_{\mathrm{uav}}}\right), \quad u \neq v,
\]
where $r_{\mathrm{uav}}$ denotes the homogeneous UAV communication range in Eq.~\eqref{eq:comm_range}.
The fleet-averaged separation is
\begin{equation}
\xi^{(t)} = \operatorname{mean}_{u \neq v} \zeta_{uv}^{(t)},
\label{eq:uav_separation}
\end{equation}
where the mean is taken over all distinct relay pairs with $u,v\in\{0,\ldots,N_{\mathrm{uav}}{-}1\}$ and $u\neq v$.
We further define the heading-switch ratio $\kappa^{(t)}\in[0,1]$ as the fraction of UAVs satisfying $m_u^{(t)}\neq m_u^{(t-1)}$, with $\kappa^{(0)}{=}0$, which quantifies the extent of fleet-wide heading action oscillations.

Within each step, delivery stress is defined strictly on ground nodes $j<N_{\mathrm{gr}}$, and only connected pairs satisfying $C_{i_u j}^{(t)}{=}1$ contribute to relay-local congestion fields.

\textbf{2) Ground-Node Delivery Stress:}
Each ground vehicle $j$ is assigned a scalar delivery stress to measure message backlog urgency:
\begin{equation}
\sigma_j^{(t)} = b_j^{(t)} \left(1 + \bar{\upsilon}_j^{(t)}\right),
\label{eq:delivery_stress}
\end{equation}
where $b_j^{(t)}$ denotes buffer utilization in Eq.~\eqref{eq:buf_util} and $\bar{\upsilon}_j^{(t)}$ denotes the average TTL urgency factor
\begin{equation}
\bar{\upsilon}_j^{(t)} = \frac{1}{|\mathcal{M}_j|} \sum_{q \in \mathcal{M}_j} \max\left(0,\, 1 - \frac{\tau_q}{T_0}\right).
\label{eq:ttl_urgency}
\end{equation}
Here, $\mathcal{M}_j$ denotes undelivered messages at ground node $j$, $\tau_q$ is the remaining TTL of packet $q$, and $T_0$ is the global maximum TTL reference.
We set $\sigma_{i_u}^{(t)}{=}0$ for all UAV nodes, which means UAVs do not participate in the calculation of delivery stress.

\textbf{3) In-Range Congestion Density and Sector-Wise Congestion:}
For relay $u$ mapped to $i_u$, only contact-valid ground nodes contribute to the in-range congestion density and sector-wise congestion.
We summarize in-range delivery stress for each relay, and average it over the fleet as
\begin{align}
\rho_{i_u}^{(t)} &= \min\left(1,\, \frac{1}{2N_{\mathrm{gr}}}\sum_{j=0}^{N_{\mathrm{gr}}-1} C_{i_u j}^{(t)} \sigma_j^{(t)}\right), \nonumber \\
\rho^{(t)} &= \frac{1}{N_{\mathrm{uav}}} \sum_{u=0}^{N_{\mathrm{uav}}-1} \rho_{i_u}^{(t)}.
\label{eq:uav_stress_density}
\end{align}

To obtain sector-wise congestion, delivery-stress values are aggregated into eight compass heading bins.
Let $\hat{\mathbf{r}}_{i_u j}^{(t)}$ denote the unit vector from relay $i_u$ to ground node $j$,
\[
\hat{\mathbf{r}}_{i_u j}^{(t)} = \frac{\mathbf{p}_j^{(t)} - \mathbf{p}_{i_u}^{(t)}}{\left\| \mathbf{p}_j^{(t)} - \mathbf{p}_{i_u}^{(t)} \right\|_2},
\]
and let ${k_{i_u j}^{*}}^{(t)}=\arg\max_{k'} \bigl( \hat{\mathbf{r}}_{i_u j}^{(t)\top} \mathbf{d}_{k'} \bigr)$ denote the heading bin best aligned with node $j$.
The sector congestion in bin \(k\) for relay \(u\) sums the stress of contacts whose best-aligned bin is \(k\),
\begin{equation}
S_{u,k}^{(t)} = \min\left(1,\, \frac{1}{2N_{\mathrm{gr}}}\sum_{j=0}^{N_{\mathrm{gr}}-1} C_{i_u j}^{(t)} \sigma_j^{(t)} \cdot \mathbb{I}\left[k = {k_{i_u j}^{*}}^{(t)}\right]\right),
\label{eq:sector_mass}
\end{equation}
where $\mathbb{I}(\cdot)$ is the indicator function.
Scalars $S_{u,k}^{(t)}$ are stacked into the sector congestion vector $\mathbf{S}_u^{(t)}=\bigl[S_{u,0}^{(t)},\ldots,S_{u,7}^{(t)}\bigr]$ (sector stress) around relay $u$.

\textbf{4) UAV Navigation Observation Features:}
\label{sec:system:uav_nav}
Let $K_{\sigma}\ge 0$ be a hyperparameter. The scenario occupies a rectangle of width $W$ and height $H$, and distances that enter routing features are normalized by the world span $L_{\mathrm{w}}=\max(W,H)$. The clipping operator is $\mathrm{clip}(v,a,b)=\max\bigl(a,\min(b,v)\bigr)$, and $\varepsilon>0$ is a small numerical constant.
Define $\mathcal{J}_{\sigma}^{(t)} = \{j_1,\ldots,j_{K_{\sigma}}\}$ as the set of ground nodes with the largest delivery‑stress values $\sigma_j^{(t)}$ (Eq.~\eqref{eq:delivery_stress}), sorted in descending order.
For each $j_r \in \mathcal{J}_{\sigma}^{(t)}$, we define the clipped normalized coordinate offsets
\begin{align*}
\tilde{\Delta}x_{u,j_r}^{(t)} &= \mathrm{clip}\bigl((p_{j_r,x}^{(t)} - p_{i_u,x}^{(t)}) / W, -1, 1\bigr),\\
\tilde{\Delta}y_{u,j_r}^{(t)} &= \mathrm{clip}\bigl((p_{j_r,y}^{(t)} - p_{i_u,y}^{(t)}) / H, -1, 1\bigr),
\end{align*}
which are computed from the planar coordinates specified in Eq.~\eqref{eq:distance}.
Stacking these offset components yields
\begin{equation}
\mathbf{e}_u^{(t)} = \bigl[\tilde{\Delta}x_{u,j_1}^{(t)},\, \tilde{\Delta}y_{u,j_1}^{(t)},\, \ldots,\, \tilde{\Delta}x_{u,j_{K_{\sigma}}}^{(t)},\, \tilde{\Delta}y_{u,j_{K_{\sigma}}}^{(t)}\bigr]^{\top}.
\label{eq:e_u_offset}
\end{equation}

Consider the set of in‑contact ground nodes $\mathcal{N}_u^{(t)} = \{j < N_{\mathrm{gr}} : C_{i_u j}^{(t)} = 1\}$.
We first calculate the stress‑weighted spatial centroid
\begin{equation}
\mathbf{c}_u^{(t)} = \frac{\sum_{j\in\mathcal{N}_u^{(t)}} \sigma_j^{(t)}\, \mathbf{p}_j^{(t)}}
{\max\!\bigl(\varepsilon,\, \sum_{j\in\mathcal{N}_u^{(t)}} \sigma_j^{(t)}\bigr)},
\label{eq:n_u_centroid}
\end{equation}
and then obtain the intermediate displacement $\boldsymbol{\delta}_u^{(t)}=\mathbf{c}_u^{(t)}-\mathbf{p}_{i_u}^{(t)}$.
The corresponding unit direction vector $\hat{\mathbf{n}}_u^{(t)}$ is given by
\begin{equation}
\hat{\mathbf{n}}_u^{(t)}=
\begin{cases}
\boldsymbol{\delta}_u^{(t)}/\bigl\|\boldsymbol{\delta}_u^{(t)}\bigr\|_2, & \bigl\|\boldsymbol{\delta}_u^{(t)}\bigr\|_2>\varepsilon;\\[4pt]
\mathbf{0}, & \text{otherwise},
\end{cases}
\label{eq:n_u_direction}
\end{equation}

By assembling the aforementioned three feature blocks, we obtain
\begin{equation}
\mathbf{z}_u^{(t)} = \bigl[\mathbf{S}_u^{(t)};\, \mathbf{e}_u^{(t)};\, \hat{\mathbf{n}}_u^{(t)}\bigr],
\qquad
D_y = 8 + 2K_{\sigma} + 2.
\label{eq:hotspot_block_dim}
\end{equation}

The \emph{navigation vector} $\mathbf{z}_u^{(t)} \in \mathbb{R}^{D_y}$ is a fixed‑dimensional local feature block fed into the UAV heading policy (Eq.~\eqref{eq:uav_motion}).
It concatenates the sector‑stress feature $\mathbf{S}_u^{(t)}$ (Eq.~\eqref{eq:sector_mass}), the Top‑$K_{\sigma}$ offset block $\mathbf{e}_u^{(t)} \in \mathbb{R}^{2K_{\sigma}}$, and the unit centroid‑direction vector $\hat{\mathbf{n}}_u^{(t)} \in \mathbb{R}^{2}$.

\section{Problem Formulation}
\label{sec:problem}
\label{sec:mdp}

In this section, we outline the optimization goals of the Joint UAV Mobility and DTN Routing Framework, cast SCF control as a finite-horizon cooperative sequential decision process under factored policies, and define the observation sets together with a per-step surrogate of the episodic objective.

\subsection{Objective Equation}
\label{sec:problem:objectives}
\label{sec:problem:p1}

We aim to maximize the successful message delivery of SCF messages while limiting TTL expiry, buffer-induced loss, and uncontrolled UAV congestion, and furthermore to shape UAV headings and fleet positioning so that the relay fleet remains dispersed and headings remain stable.
At each step $t$, we define $D^{(t)}$, $E^{(t)}$ and $R^{(t)}$ as the respective counts of delivered, expired and dropped messages, and $\bar{B}^{(t)}{=}\frac{1}{N}\sum_{i=0}^{N-1}b_i^{(t)}$ as the mean buffer utilization (Eq.~\eqref{eq:buf_util}).
We treat successful delivery as the primary criterion and the loss counts as penalties, and formulate the delivery objective as
\begin{equation}
\begin{aligned}
J_{\mathrm{del}}(\pi)
&= \mathbb{E}_{\pi}\!\Biggl[\sum_{t=0}^{T_{\max}}
\bigl(
\alpha_d D^{(t)} + \alpha_e E^{(t)} + \alpha_{dr} R^{(t)}
\\
&\qquad
+ \alpha_b \bar{B}^{(t)}
\bigr)\Biggr],
\end{aligned}
\label{eq:p1_delivery}
\end{equation}
where $\alpha_d>0$ rewards delivered messages, and $\alpha_e,\alpha_{dr},\alpha_b\le 0$ penalize TTL expiry, buffer drops, and mean congestion, respectively.

The separation and heading-switch terms below are scaled by gates. Once a step records a delivery, $g_s$ reduces the separation incentive and $g_m$ reduces the heading-oscillation penalty:
\begin{equation}
\begin{aligned}
g_s\!\bigl(D^{(t)}\bigr) &=
\begin{cases}
1, & D^{(t)}=0,\\
0.3, & \text{otherwise,}
\end{cases}
\\[2pt]
g_m\!\bigl(D^{(t)}\bigr) &=
\begin{cases}
1, & D^{(t)}=0,\\
0.5, & \text{otherwise.}
\end{cases}
\end{aligned}
\label{eq:gates}
\end{equation}
The corresponding mobility objective is
\begin{equation}
\begin{aligned}
J_{\mathrm{uav}}(\pi)
&= \mathbb{E}_{\pi}\!\Biggl[\sum_{t=0}^{T_{\max}}
\Bigl(
\alpha_{\rho}\,\rho^{(t)} + \alpha_{c}\,\rho^{(t)} D^{(t)}
+ \alpha_x\, g_s\!\bigl(D^{(t)}\bigr)\,\xi^{(t)}
\\
&\qquad\qquad
- \alpha_k\, g_m\!\bigl(D^{(t)}\bigr)\,\kappa^{(t)}
\Bigr)\Biggr].
\end{aligned}
\label{eq:p1_uav}
\end{equation}
Here \(\alpha_\rho\) rewards the in-range congestion density \(\rho^{(t)}\) (Eq.~\eqref{eq:uav_stress_density}), \(\alpha_c\) couples that density to step deliveries, \(\alpha_x\) encourages the fleet separation \(\xi^{(t)}\) (Eq.~\eqref{eq:uav_separation}), and \(\alpha_k\) penalizes the heading-switch ratio \(\kappa^{(t)}\) (Sec.~\ref{sec:system:uav_dispersion}).
Although merely optimizing the delivery objective \(J_{\mathrm{del}}(\pi)\) suffers from sparse, delayed step rewards caused by the SCF transmission paradigm, optimizing only the UAV mobility objective \(J_{\mathrm{uav}}(\pi)\) may drive relays to gather around congestion hotspots without increasing delivered messages. To avoid the drawbacks of single-objective optimization, we integrate the two objectives into a unified cooperative control objective as
\begin{equation}
\max_{\pi}\; J(\pi),
\qquad
J(\pi) = J_{\mathrm{del}}(\pi) + J_{\mathrm{uav}}(\pi),
\label{eq:p1}
\end{equation}
subject to the SCF and action rules of Sec.~\ref{sec:system}, in which ground mobility is determined by predefined rules, whereas UAV motion and routing decisions are determined by $\pi_\theta$.

\subsection{Joint Decision Process and Factored Policy}
\label{sec:problem:sdp}
\label{sec:problem:joint_coupling}

\paragraph{Sequential decision process.}
We recast the system model of Sec.~\ref{sec:system} as a cooperative sequential decision process whose per-step transition follows the step-wise execution pipeline in Fig.~\ref{fig:system_model}(b). That pipeline executes UAV heading actions, opportunistic replication, and network-state refresh in that order, then returns decentralized observations $\mathbf{o}^{(t)}$, until the episode terminates at $t{=}T_{\max}$. Let $x^{(t)}$ denote the full environment state: GNSS-localized positions $\{\mathbf{p}_i^{(t)}\}$, buffers $\{\mathcal{B}_i^{(t)}\}$, contact matrix $\mathbf{C}^{(t)}$, and the in-range congestion density and sector-wise congestion metrics of Sec.~\ref{sec:system:stress}.
The joint action is
\begin{equation}
\mathbf{a}^{(t)}=\bigl(a_0^{(t)},\ldots,a_{N-1}^{(t)},\,m_0^{(t)},\ldots,m_{N_{\mathrm{uav}}-1}^{(t)}\bigr),
\label{eq:joint_action}
\end{equation}
where $a_i^{(t)}\in\mathcal{A}_R$ selects a routing candidate or idle action and $m_u^{(t)}\in\mathcal{A}_M$ selects a UAV heading bin (Eq.~\eqref{eq:uav_motion}).
To enforce motion-before-routing within each slot, the one-step transition is
\begin{equation}
x^{(t+1)}
=
\Phi\!\Big(
  \Psi\big(x^{(t)},\,\{m_u^{(t)}\}_{u=0}^{N_{\mathrm{uav}}-1}\big),\,
  \{a_i^{(t)}\}_{i=0}^{N-1}
\Big),
\label{eq:coupled_transition}
\end{equation}
where $\Psi$ updates relay positions from the heading actions, and $\Phi$ then completes replication, TTL decay, and contact recomputation.

\paragraph{Structural couplings.}
Two bidirectional structural couplings exist between UAV mobility and routing subsystems, as detailed below.

\textbf{(C1) Topology-to-routing coupling}
For any node $i$, the reachable forwarding neighbors are $\{j:C_{ij}^{(t)}=1\}$, where $j$ indexes a node in contact with $i$. Contact availability relies on pairwise Euclidean distances $d_{ij}^{(t)}$, which are directly influenced by UAV positions.
When a relay $u$ adjusts its heading $m_u^{(t)}$ to move toward congested ground zones, it can establish new communication links that were unavailable in the previous time step. This creates extra forwarding opportunities that routing strategies cannot access under a static, fixed network topology.

\textbf{(C2) Congestion-to-mobility coupling}
Message replication and successful deliveries modify node buffer occupancy and the spatial distribution of pending packets. These buffer updates enter ground-node delivery stress $\sigma_j^{(t)}$, which shapes the in-range congestion density and sector-wise congestion of relays in contact with node $j$, thereby guiding the heading actions selected in subsequent steps.

The state transition operators $\Psi$ (UAV mobility update) and $\Phi$ (routing \& network refresh) follow a rigid execution order: UAV positions are updated before any message replication within one time step. The two operators cannot be swapped, which introduces sequential cross-coupling between subsystems. As a result, we cannot optimize UAV mobility and routing via separate, decoupled iterative loops.

\label{sec:problem:action}
\paragraph{Factored policy and action spaces.}
Instead of learning a single joint policy for global network flow routing, we factor the control task into \(N+N_{\mathrm{uav}}\) cooperative decision units, in which each node~$i$ has one routing unit and each UAV~$u$ has one motion unit.

Since each UAV is indexed both as graph node $i_u$ and as relay $u$, Eq.~\eqref{eq:joint_action} includes its routing action $a_{i_u}^{(t)}$ together with its heading action $m_u^{(t)}$. These actions are chosen from distinct observations, $o_{i_u}^{(t)}$ for routing and ${o_u^{\mathrm{M}}}^{(t)}$ for heading, where the superscript $\mathrm{M}$ marks the heading factor.

A relay must both reach contacts through motion and use them for SCF replication. Each node selects at most one local transfer from a capped candidate set ($|\mathcal{A}_R|{=}K{+}1$), and each UAV selects one of $B{=}8$ headings.
Let $\mathcal{F}_i^{(t)}\subseteq\mathcal{A}_R$ denote the routing actions feasible at node~$i$. The number of joint actions at step~$t$ is
\begin{equation}
|\mathcal{A}_{\mathrm{joint}}^{(t)}| \;=\;
\prod_{i=0}^{N-1} |\mathcal{F}_i^{(t)}| \;\times\; B^{N_{\mathrm{uav}}},
\label{eq:factored_action_size}
\end{equation}
where $|\mathcal{F}_i^{(t)}|$ depends on $\mathbf{C}^{(t)}$ and therefore depends on prior motion, while each factor conditions only on local observations. Given decentralized observations $\mathbf{o}^{(t)}$, node~$i$ selects $a_i^{(t)}\in\mathcal{A}_R$ according to $\pi_{\theta,i}(a_i^{(t)}\mid o_i^{(t)})$, and relay~$u$ selects $m_u^{(t)}\in\mathcal{A}_M$ according to $\pi_{\theta,u}^{\mathrm{M}}(m_u^{(t)}\mid {o_u^{\mathrm{M}}}^{(t)})$.
This specializes the factored policy of Sec.~\ref{sec:related:prelim} by setting
\begin{equation}
M=N+N_{\mathrm{uav}}=N_{\mathrm{gr}}+2N_{\mathrm{uav}},
\label{eq:decision_unit_count}
\end{equation}
These \(M\) units are all the action units.
Each of the \(N\) graph nodes contributes one routing factor for SCF forwarding, and each UAV additionally contributes a heading factor for aerial motion:
\begin{align}
\pi_\theta(\mathbf{a}^{(t)}\mid\mathbf{o}^{(t)}) &=
\prod_{i=0}^{N-1}\pi_{\theta,i}\!\left(a_i^{(t)}\mid o_i^{(t)}\right) \nonumber\\
&\quad \times \prod_{u=0}^{N_{\mathrm{uav}}-1}\pi_{\theta,u}^{\mathrm{M}}\!\left(m_u^{(t)}\mid {o_u^{\mathrm{M}}}^{(t)}\right).
\label{eq:policy_factorization}
\end{align}
Every factor in the product uses the same parameter vector $\theta$.

\paragraph{Per-step decomposition.}
\label{sec:problem:decomposition}
The per-step summand of \(J_{\mathrm{del}}\) in Eq.~\eqref{eq:p1_delivery} and the per-step summand of \(J_{\mathrm{uav}}\) in Eq.~\eqref{eq:p1_uav} together define the shared team reward
\begin{equation}
\begin{aligned}
r^{(t)}
&=
\alpha_d D^{(t)} + \alpha_e E^{(t)} + \alpha_{dr} R^{(t)}\\
&\quad
+ \alpha_b \bar{B}^{(t)}\\
&\quad
+ \alpha_{\rho}\,\rho^{(t)} + \alpha_{c}\,\rho^{(t)}\, D^{(t)}\\
&\quad
+ \alpha_x\, g_s\!\bigl(D^{(t)}\bigr)\,\xi^{(t)}
- \alpha_k\, g_m\!\bigl(D^{(t)}\bigr)\,\kappa^{(t)},
\end{aligned}
\label{eq:team_reward}
\end{equation}
which is assigned to all \(N{+}N_{\mathrm{uav}}\) nodes. Thus the objective \(J(\pi)\) in Eq.~\eqref{eq:p1} is the expected sum of \(r^{(t)}\).
Learning retains the factorization in Eq.~\eqref{eq:policy_factorization} and maximizes the discounted return
\begin{equation}
J(\theta)=\mathbb{E}_{\pi_\theta}\!\left[\sum_{t=0}^{T_{\max}} \gamma^{t}\, r^{(t)}\right],
\label{eq:p1_decomposition}
\end{equation}
Here \(J(\theta)\) is the objective \(J\) evaluated at the parameterized policy \(\pi_\theta\), and \(\gamma\in(0,1]\) is the discount factor. When \(\gamma=1\), Eq.~\eqref{eq:p1_decomposition} is equivalent to the undiscounted finite-horizon objective \(J(\pi)\) in Eq.~\eqref{eq:p1}.

\subsection{Observation Model}
\label{sec:problem:obs}
\label{sec:problem:marl}

Directly maximizing \(J(\pi)\) in Eq.~\eqref{eq:p1} is computationally intractable under contact-limited sensing.
We therefore define the observation sets of the factored decision units.
These units form a fully cooperative, partially observable multi-agent team (cf.\ Sec.~\ref{sec:related:prelim}). Because sensing is contact-limited, a routing unit observes neighbor statistics retained from the last contact and cached destination estimates when destinations are out of contact, whereas a UAV heading unit observes local navigation features together with a summary of network-wide or in-contact statistics.

\label{sec:problem:exchange}
\paragraph{Neighbor exchange summaries.}
Each routing node $i$ observes its own $b_i^{(t)}$ and $\bar{\nu}_i^{(t)}$ (Eqs.~\eqref{eq:buf_util} and~\eqref{eq:norm_degree}) from on-board statistics, but cannot read other nodes' current buffer utilization or contact degree unless they meet.
To support contact-limited routing features, each node $i$ maintains a neighbor exchange table.
Entry $(\tilde{b}_{ij}^{(t)},\tilde{\bar{\nu}}_{ij}^{(t)})$ stores $i$'s last-exchange estimates of neighbor $j$'s live buffer utilization $b_j^{(t)}$ and normalized contact degree $\bar{\nu}_j^{(t)}$, and is refreshed to $(b_j^{(t)},\bar{\nu}_j^{(t)})$ at every step with $C_{ij}^{(t)}{=}1$.
Between meetings, these estimates remain the values of $b_j$ and $\bar{\nu}_j$ observed at the last exchange.
Let $\iota_{ij}^{\mathrm{ex}\,(t)}\ge 0$ denote the steps elapsed since that exchange, and let \(\omega_{\mathrm{ex}}>0\) denote the exchange decay factor that discounts older estimates. The exchange-weighted neighbor buffer at node~$i$ is
\begin{equation}
\bar{b}_i^{\mathrm{nb}\,(t)}
=
\frac{
\sum_{j:C_{ij}^{(t)}{=}1}\exp(-\omega_{\mathrm{ex}}\iota_{ij}^{\mathrm{ex}\,(t)})\,\tilde{b}_{ij}^{(t)}
}{
\sum_{j:C_{ij}^{(t)}{=}1}\exp(-\omega_{\mathrm{ex}}\iota_{ij}^{\mathrm{ex}\,(t)})
}.
\label{eq:neighbor_buf_mean}
\end{equation}
This average equals \(0\) when node~$i$ has no neighbor in contact.

The local routing state \(\mathbf{x}_i^{(t)}\in\mathbb{R}^{8}\) concatenates node~$i$'s own buffer and traffic statistics, its speed ratio and normalized position, its normalized contact degree, and the neighbor-buffer summary of Eq.~\eqref{eq:neighbor_buf_mean}.
Let \(\beta_i^{\mathrm{g}\,(t)}\) and \(\beta_i^{\mathrm{d}\,(t)}\) denote the fractions of all messages created so far that were generated and delivered at node~$i$, respectively.
The speed entry is \(\bar{v}_i^{(t)}=\|\mathbf{v}_i^{(t)}\|_2/v_{\mathrm{gr}}\). With every ground vehicle at constant speed \(v_{\mathrm{gr}}\) and every UAV relay at constant speed \(v_{\mathrm{uav}}\) (Sec.~\ref{sec:system:model}), this entry equals \(1\) for a ground node and \(v_{\mathrm{uav}}/v_{\mathrm{gr}}\) for a UAV relay. Both node types use the divisor \(v_{\mathrm{gr}}\), so the feature occupies the same slot.
The normalized position is \(\bar{\mathbf{p}}_i^{(t)}=(p_{i,x}^{(t)}/W,\,p_{i,y}^{(t)}/H)\), and the normalized contact degree \(\bar{\nu}_i^{(t)}\) follows Eq.~\eqref{eq:norm_degree}. Accordingly,
\begin{equation}
\begin{aligned}
\mathbf{x}_i^{(t)}
&=
\bigl[
b_i^{(t)},\,
\beta_i^{\mathrm{g}\,(t)},\,
\beta_i^{\mathrm{d}\,(t)},\,
\bar{v}_i^{(t)},\,
\bar{\nu}_i^{(t)},\\
&\qquad
p_{i,x}^{(t)}/W,\,
p_{i,y}^{(t)}/H,\,
\bar{b}_i^{\mathrm{nb}\,(t)}
\bigr]^{\top}.
\end{aligned}
\label{eq:local_state}
\end{equation}
The leading block \(\mathbf{f}_i^{(t)}=(b_i^{(t)},\beta_i^{\mathrm{g}\,(t)},\beta_i^{\mathrm{d}\,(t)},\bar{v}_i^{(t)},\bar{\nu}_i^{(t)})\) is retained for the UAV context features defined later in this subsection.

Routing further conditions on up to \(K\) forwarding candidates. For slot \(\ell\in\{1,\ldots,K\}\) that carries message \(q\) toward destination \(d\), the candidate feature is
\begin{equation}
\begin{aligned}
\mathbf{c}_{i,\ell}^{(t)}
&=
\bigl(
b_{id}^{\mathrm{dst}\,(t)},\,
\tilde{\bar{\nu}}_{id}^{(t)},\,
\tau_q/T_0,\,
\iota_q^{(t)}/T_0,\\
&\qquad
h_q^{(t)}/(N-1),\,
\varsigma_q^{(t)},\,
d_{id}^{(t)}/L_{\mathrm{w}}
\bigr)
\in\mathbb{R}^{7}.
\end{aligned}
\label{eq:candidate_feat}
\end{equation}
The first two entries describe the destination: \(b_{id}^{\mathrm{dst}\,(t)}\) equals the live buffer \(b_d^{(t)}\) if \(C_{id}^{(t)}{=}1\) and the exchange estimate \(\tilde{b}_{id}^{(t)}\) otherwise, whereas \(\tilde{\bar{\nu}}_{id}^{(t)}\) always stores the exchange estimate of \(\bar{\nu}_d^{(t)}\).
The remaining entries describe the message and the geometric hop: remaining TTL \(\tau_q\) and age \(\iota_q^{(t)}\) normalized by \(T_0\), hop count \(h_q^{(t)}\) normalized by \(N-1\), normalized payload \(\varsigma_q^{(t)}=s_q/S_{\max}\), and distance \(d_{id}^{(t)}/L_{\mathrm{w}}\).
When fewer than \(K\) candidates are available, unused slots are padded to keep a fixed observation width, and the corresponding actions are excluded from \(\mathcal{F}_i^{(t)}\).
Collecting the local node state \(\mathbf{x}_i^{(t)}\) and the candidate feature vectors \(\{\mathbf{c}_{i,\ell}^{(t)}\}_{\ell=1}^{K}\) yields the routing observation
\begin{equation}
o_i^{(t)}=\bigl(\mathbf{x}_i^{(t)},\,\{\mathbf{c}_{i,\ell}^{(t)}\}_{\ell=1}^{K}\bigr).
\label{eq:routing_obs}
\end{equation}

\label{sec:problem:uav_obs}
\label{sec:mdp:uav_obs}
\paragraph{UAV motion observations and context summaries.}
Each UAV perceives congestion hotspots indirectly through the navigation features \(\mathbf{z}_u^{(t)}\in\mathbb{R}^{D_y}\) of Sec.~\ref{sec:system:uav_nav} (Eq.~\eqref{eq:hotspot_block_dim}), and through either the network-wide summary \({g_u^{\mathrm{G}}}^{(t)}\), which is shared across the fleet, or the contact-limited summary \({g_u^{\mathrm{L}}}^{(t)}\), which is limited to the ground nodes in contact with node \(i_u\).
Let \(q_{\mathrm{p}}^{(t)}\) be the fraction of messages created so far that are still pending, excluding those already delivered, expired, or dropped. The normalized time \(\tilde{t}^{(t)}=t/T_{\max}\) is the fraction of the episode horizon elapsed by step \(t\). Both scalars are included only in the network-wide summary \({g_u^{\mathrm{G}}}^{(t)}\) and are excluded from the contact-limited summary \({g_u^{\mathrm{L}}}^{(t)}\).
To give every relay a shared episode-level context during training, we pack network-wide buffer, position, and contact statistics into
\begin{equation}
\begin{aligned}
{g_u^{\mathrm{G}}}^{(t)}
&=
\Bigl(
\operatorname{mean}_{i=0,\ldots,N-1}\!\bigl[\mathbf{f}_i^{(t)};\bar{\mathbf{p}}_i^{(t)}\bigr],\\
&\qquad
\operatorname{std}_{i=0,\ldots,N-1}\!\bigl[\mathbf{f}_i^{(t)};\bar{\mathbf{p}}_i^{(t)}\bigr],\\
&\qquad
\frac{1}{N^{2}}\sum_{i,j=0}^{N-1}C_{ij}^{(t)},\;
\operatorname{mean}_{i=0,\ldots,N-1}\bar{\nu}_i^{(t)},\\
&\qquad
\operatorname{std}_{i=0,\ldots,N-1}\bar{\nu}_i^{(t)},\;
q_{\mathrm{p}}^{(t)},\;
\tilde{t}^{(t)}
\Bigr),
\end{aligned}
\label{eq:global_summary}
\end{equation}
where \(i\) and \(j\) are the node indices of Sec.~\ref{sec:system:model}, and \(1/N^{2}\) averages all \(N^{2}\) entries of \(\mathbf{C}^{(t)}\), including the zero diagonal.
The contact-limited summary uses only ground nodes currently in contact with \(i_u\),
\begin{equation}
\begin{aligned}
{g_u^{\mathrm{L}}}^{(t)}
&=
\Bigl(
\operatorname{mean}_{j<N_{\mathrm{gr}},\,C_{i_u j}^{(t)}=1}\!\bigl[\mathbf{f}_j^{(t)};\bar{\mathbf{p}}_j^{(t)}\bigr],\\
&\qquad
\operatorname{std}_{j<N_{\mathrm{gr}},\,C_{i_u j}^{(t)}=1}\!\bigl[\mathbf{f}_j^{(t)};\bar{\mathbf{p}}_j^{(t)}\bigr]
\Bigr),
\end{aligned}
\label{eq:local_summary}
\end{equation}
where \(i_u\) is the graph-node index of relay \(u\), and \(j\) runs over ground nodes in contact with \(i_u\). When no ground node is in contact with \(i_u\), the mean and the standard deviation are both zero.
UAV relay \(u\) receives the observation
\begin{equation}
{o_u^{\mathrm{M}}}^{(t)}=\bigl(\mathbf{z}_u^{(t)},\,g_u^{(t)}\bigr),
\quad
g_u^{(t)}\in\bigl\{{g_u^{\mathrm{G}}}^{(t)},\,{g_u^{\mathrm{L}}}^{(t)}\bigr\},
\label{eq:uav_obs}
\end{equation}
where $\mathbf{z}_u^{(t)}=\bigl[\mathbf{S}_u^{(t)};\,\mathbf{e}_u^{(t)};\,\hat{\mathbf{n}}_u^{(t)}\bigr]$ stacks the sector-wise congestion, Top-$K_{\sigma}$ offset, and centroid-direction blocks specified in Sec.~\ref{sec:system:uav_nav}.
Next-hop forwarding at UAV relay \(i_u\) uses the routing observation \(o_{i_u}^{(t)}\) of Eq.~\eqref{eq:routing_obs}, which has the same structure as the observation of a ground vehicle.

\section{Proposed \texorpdfstring{\Judor{}}{JUDOR} Algorithm}
\label{sec:method}
\begin{figure}[!t]
\centering
\includegraphics[width=\columnwidth]{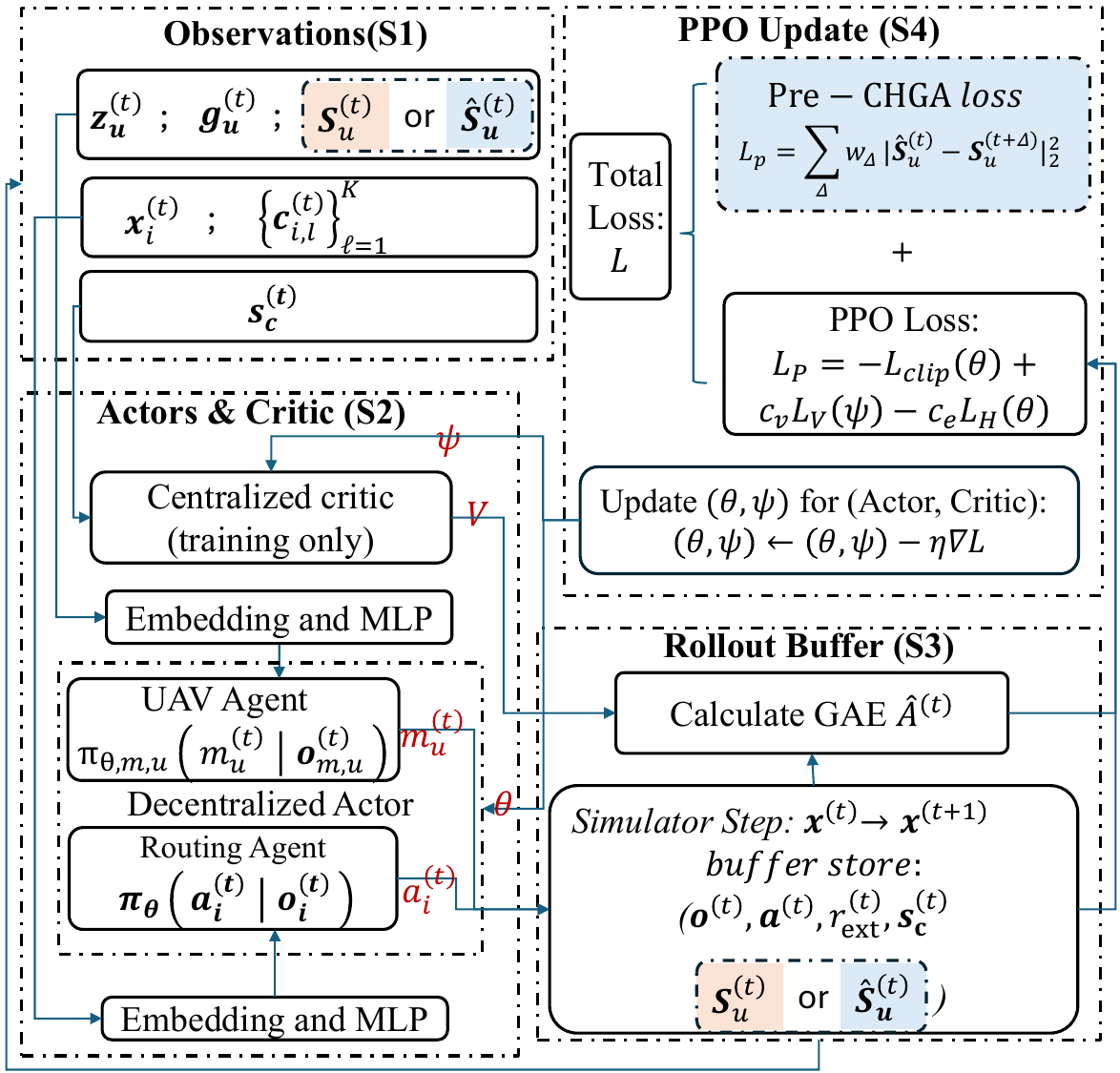}
\captionsetup{justification=justified}
\caption{JUDOR CTDE--PPO architecture for one training epoch.
Red denotes GT-CHGA and blue denotes Pred-CHGA. ``\(+\)'' denotes feature concatenation.
\textbf{(S1)}~\((\mathbf{z}_u^{(t)}\), \(g_u^{(t)})\) (Eqs.~\eqref{eq:uav_obs}, \eqref{eq:hotspot_block_dim}, \eqref{eq:global_summary}--\eqref{eq:local_summary}); dashed \(\mathbf{S}_u^{(t)}\) or \(\hat{\mathbf{S}}_u^{(t)}\) (Eqs.~\eqref{eq:sector_mass}, \eqref{eq:lstm_predictor}); \((\mathbf{x}_i^{(t)}\), \(\{\mathbf{c}_{i,\ell}^{(t)}\}_{\ell=1}^{K})\) (Eqs.~\eqref{eq:routing_obs}, \eqref{eq:local_state}--\eqref{eq:candidate_feat}); \(s_{\mathrm{c}}^{(t)}\) (Eq.~\eqref{eq:critic_summary}).
\textbf{(S2)}~decentralized actors and the training-only critic; \textbf{(S3)}~\((s_{\mathrm{c}}^{(t)}\), \(\mathbf{o}^{(t)}\), \(\mathbf{a}^{(t)}\), \(r_{\mathrm{ext}}^{(t)},\ldots)\) (Eq.~\eqref{eq:reward_full}); \textbf{(S4)}~\(L_P\) (Eq.~\eqref{eq:ppo_base}) and optional \(L_p\) (Eq.~\eqref{eq:aux_loss}).}
\label{fig:system_arch}
\end{figure}

Based on the system model and factored POMDP established in Secs.~\ref{sec:system} and~\ref{sec:problem}, this section introduces the proposed \Judor{} algorithm under the CTDE multi‑agent reinforcement‑learning framework.
Fig.~\ref{fig:system_arch} illustrates the overall training pipeline of \Judor{} within a single training epoch.
Sec.~\ref{sec:method:reward} defines optional congestion-hotspot-guided alignment (CHGA), which aligns UAV headings to either ground-truth sector congestion under GT-CHGA or the LSTM forecast sector congestion under Pred-CHGA.
Sec.~\ref{sec:method:ctde} instantiates the Actors \& Critic block~(S2), which consumes the Observations block~(S1).
Sec.~\ref{sec:method:training} details the Rollout Buffer block~(S3) and the PPO Update block~(S4).

\label{sec:method:overview}

\subsection{Congestion-Hotspot-Guided Alignment}
\label{sec:method:reward}
\label{sec:problem:reward}
\label{sec:mdp:reward}

The per-step summand of the objective \(J(\pi)\) in Eq.~\eqref{eq:p1} is the team reward \(r^{(t)}\) in Eq.~\eqref{eq:team_reward}, which rewards deliveries and penalizes expiry, drops, buffer pressure, and unfavorable fleet geometry.

Under the SCF paradigm, however, \(D^{(t)}\) is sparse, so \(r^{(t)}\) supplies insufficient learning guidance for the UAV heading action \(m_u^{(t)}\).
CHGA mitigates this shortfall.
CHGA quantifies the alignment between each UAV’s executed heading and the sector‑wise congestion metric defined in Sec.~\ref{sec:system:stress}.

Let $\boldsymbol{\omega}=(\omega_0,\ldots,\omega_{B-1})\in\mathbb{R}_{\ge 0}^{B}$ denote a non‑negative sector weight vector defined over the $B=8$ discrete heading bins introduced in Sec.~\ref{sec:system:control}.
To aggregate the sector‑wise weight distribution into a single reference unit direction for evaluating UAV heading choices, we define the reference heading vector as
\begin{equation}
\mathbf{d}_{\mathrm{ref}}(\boldsymbol{\omega})=
\frac{\sum_{k=0}^{B-1} \omega_k\,\mathbf{d}_k}{\left\lVert\sum_{k=0}^{B-1} \omega_k\,\mathbf{d}_k\right\rVert_2},
\label{eq:ref_heading}
\end{equation}
Here $\mathbf{d}_k$ stands for the time‑invariant unit direction vector corresponding to heading bin $k$, while $\mathbf{d}_{m_u^{(t)}}$ is the direction vector selected by relay $u$’s executed heading action (cf. Eq.~\eqref{eq:uav_motion}).
An alignment score \(\varphi^{(t)}\) quantifies the fleet‑average directional matching between the executed UAV headings $\mathbf{d}_{m_u^{(t)}}$ and the computed reference heading $\mathbf{d}_{\mathrm{ref}}(\boldsymbol{\omega})$.

When CHGA is enabled, either the red or the blue shaded region in Fig.~\ref{fig:system_arch} is active.
GT-CHGA activates the red region and uses the ground-truth sector-congestion vector \(\mathbf{S}_u^{(t)}\) directly. Pred-CHGA activates the blue region including the dashed LSTM branch, which produces the forecast \(\hat{\mathbf{S}}_u^{(t)}\).

\paragraph{GT-CHGA.}
When GT-CHGA is selected, the sector weight is the ground-truth congestion \(\boldsymbol{\omega}=\mathbf{S}_u^{(t)}\).
The score \(\varphi_g^{(t)}\) instantiates \(\varphi^{(t)}\) as the mean alignment of each executed heading with this reference:
\begin{equation}
\varphi_g^{(t)} =
\frac{1}{N_{\mathrm{uav}}}\sum_{u=0}^{N_{\mathrm{uav}}-1}
\max\!\bigl(0,\,{\mathbf{d}_{m_u^{(t)}}}^{\top}\mathbf{d}_{\mathrm{ref}}(\mathbf{S}_u^{(t)})\bigr).
\label{eq:alpha_gt}
\end{equation}

\paragraph{Pred-CHGA.}
\label{sec:method:lstm}
When Pred-CHGA is selected, the sector weight is the forecasted congestion \(\boldsymbol{\omega}=\hat{\mathbf{S}}_u^{(t)}\).
UAV \(u\) forms this forecast from its navigation vector \(\mathbf{z}_u^{(t)}\) by an LSTM followed by a two-layer MLP \(f_p\):
\begin{equation}
\hat{\mathbf{S}}_u^{(t)}
=
f_p\!\bigl(\operatorname{LSTM}(\mathbf{z}_u^{(t)})\bigr)
\in\mathbb{R}^{B},
\label{eq:lstm_predictor}
\end{equation}
which has the same dimension as \(\mathbf{S}_u^{(t)}\).
The score \(\varphi_p^{(t)}\) instantiates \(\varphi^{(t)}\) as the mean alignment of each executed heading with this reference:
\begin{equation}
\varphi_p^{(t)} =
\frac{1}{N_{\mathrm{uav}}}\sum_{u=0}^{N_{\mathrm{uav}}-1}
\max\!\bigl(0,\,{\mathbf{d}_{m_u^{(t)}}}^{\top}\mathbf{d}_{\mathrm{ref}}(\hat{\mathbf{S}}_u^{(t)})\bigr).
\label{eq:alpha_pred}
\end{equation}

To reward UAV headings pointing toward congested sectors, the learner extends Eq.~\eqref{eq:team_reward} by an alignment score \(\varphi^{(t)}\).
Let \(I_h^{(t)}\in\{0,1\}\) be the routing-activity indicator
\begin{equation}
I_h^{(t)}=
\begin{cases}
1, & \text{if at least one replication succeeds at step } t;\\
0, & \text{otherwise}.
\end{cases}
\label{eq:hop_indicator}
\end{equation}
\begin{equation}
\begin{aligned}
r_{\mathrm{ext}}^{(t)}
&=
r^{(t)}
+ \alpha_a \varphi^{(t)}
+ \alpha_{ad}\,\varphi^{(t)}\, D^{(t)}
+ \alpha_{ar}\,\varphi^{(t)}\, I_h^{(t)},
\end{aligned}
\label{eq:reward_full}
\end{equation}
where \(\varphi^{(t)}\) is \(\varphi_g^{(t)}\) under GT-CHGA or \(\varphi_p^{(t)}\) under Pred-CHGA.
If CHGA is disabled, \(r_{\mathrm{ext}}^{(t)}\) degenerates to the original team reward \(r^{(t)}\).
The constant coefficients \(\alpha_a,\alpha_{ad},\alpha_{ar}\) weight the alignment score \(\varphi^{(t)}\), its product with the delivery count \(D^{(t)}\), and its product with the routing-activity indicator \(I_h^{(t)}\) of Eq.~\eqref{eq:hop_indicator}.
The last two products reward an aligned heading more when the same step also delivers or records a successful replication, so a heading that merely points at congestion is rewarded less than one that coincides with a useful transfer.

\subsection{CTDE Actor--Critic Instantiation}
\label{sec:method:ctde}
Building upon the factored policy from Sec.~\ref{sec:problem:sdp} and the contact‑limited observation model in Sec.~\ref{sec:problem:obs}, we implement a CTDE actor–critic architecture for the joint DTN routing and UAV heading‑control task.
In real‑world deployment, network nodes suffer intermittent communication contacts and cannot access network‑wide global states at inference time. Therefore, only decentralized actors run on physical nodes, whereas the centralized critic together with its global aggregation module are used for training only and discarded during runtime.
As illustrated in blocks (S1)--(S2) of Fig.~\ref{fig:system_arch}, our framework contains two types of actor heads: routing actors for every ground and UAV node, and dedicated UAV‑direction heads for each aerial relay.

Under the local‑observation conditional‑independence assumption for the factored policy in Eq.~\eqref{eq:policy_factorization}, the joint‑policy log‑probability decomposes additively over individual routing and UAV heading actions:
\begin{equation}
\begin{aligned}
\log \pi_\theta(\mathbf{a}^{(t)}\mid\mathbf{o}^{(t)})
&=
\sum_{i=0}^{N-1}\log \pi_{\theta,i}(a_i^{(t)}\mid o_i^{(t)})
\\
&\quad
+
\sum_{u=0}^{N_{\mathrm{uav}}-1}\log \pi_{\theta,u}^{\mathrm{M}}(m_u^{(t)}\mid {o_u^{\mathrm{M}}}^{(t)}).
\end{aligned}
\label{eq:log_prob_factorization}
\end{equation}
Each term represents the log‑probability of one local action given its local observation, with all decision units sharing the parameter vector $\theta$.
Following the GAE formulation (Sec.~\ref{sec:related:prelim}), the centralized critic yields the shared team‑level advantage $\hat{A}^{(t)}$, which enables CTDE‑style multi‑agent credit assignment for locally‑observed actor updates. This gives the gradient estimator for updating all local actors:
\begin{equation}
\nabla_\theta J(\theta)
\;\approx\;
\mathbb{E}\!\left[
\hat{A}^{(t)}\,
\nabla_\theta \log \pi_\theta(\mathbf{a}^{(t)}\mid\mathbf{o}^{(t)})
\right].
\label{eq:team_policy_grad}
\end{equation}
Eq.~\eqref{eq:team_policy_grad} approximates the gradient of the parameterized objective $J(\theta)$ from Eq.~\eqref{eq:p1_decomposition}. 
We next detail the three key components: the routing actor, UAV‑direction head, and training‑only centralized critic.

\textbf{1) Routing actor.}
Corresponding to high‑level functional view block (S2) in Fig.~\ref{fig:system_arch}, the routing actor runs for every node, including ground vehicles and UAV relays.
Its input is the routing observation \(o_i^{(t)}\) of Eq.~\eqref{eq:routing_obs}, which has two parts: the node-local state \(\mathbf{x}_i^{(t)}\in\mathbb{R}^{8}\) and up to \(K\) forwarding-candidate features \(\mathbf{c}_{i,\ell}^{(t)}\in\mathbb{R}^{7}\).
We use two MLP encoders: one embeds \(\mathbf{x}_i^{(t)}\), and the other embeds each \(\mathbf{c}_{i,\ell}^{(t)}\).
Each forwarding candidate is scored from the concatenation of the node embedding and the candidate embedding, so the decision uses both the node's state and the candidate features.
The resulting \(K{+}1\) scores (idle plus \(K\) forwards) parameterize \(\pi_{\theta,i}\) over \(\mathcal{A}_R=\{0,\ldots,K\}\).
If fewer than \(K\) candidates are present, unused slots are padded and excluded from sampling.

\textbf{2) UAV direction head.}
Also implemented within block (S2) of Fig.~\ref{fig:system_arch}, each UAV relay instantiates a direction‑head module to generate discrete heading actions $m_u^{(t)}$.
The module accepts input $\boldsymbol{o}_u^{\mathrm{M}(t)}=(\mathbf{z}_u^{(t)},g_u^{(t)})$ (Eq.~\eqref{eq:uav_obs}), where $\mathbf{z}_u^{(t)}$ is defined in Sec.~\ref{sec:system:uav_nav}, and $g_u^{(t)}$ selects either network‑wide or contact‑limited context summaries (Sec.~\ref{sec:problem:obs}).
Let $\boldsymbol{\ell}_u^{(t)}\in\mathbb{R}^{B}$ denote heading logits output by motion MLP $f_m$, from which $m_u^{(t)}$ is sampled via $\operatorname{softmax}(\boldsymbol{\ell}_u^{(t)})$.
For both the CHGA‑disabled baseline and GT‑CHGA ($\lambda=0$), we have $\boldsymbol{\ell}_u^{(t)}=f_m\!\bigl([g_u^{(t)};\mathbf{z}_u^{(t)}]\bigr)$. Under GT‑CHGA, ground‑truth sector congestion $\mathbf{S}_u^{(t)}$ contributes exclusively to the extended reward $r_{\mathrm{ext}}^{(t)}$ and is never fed into $f_m$.
For Pred‑CHGA with $\lambda>0$ (dashed LSTM branch in Fig.~\ref{fig:system_arch}), the sector‑congestion forecast from Eq.~\eqref{eq:lstm_predictor} is concatenated into the MLP input:
\begin{equation}
\boldsymbol{\ell}_u^{(t)} = f_m\!\bigl([g_u^{(t)};\mathbf{z}_u^{(t)};\hat{\mathbf{S}}_u^{(t)}]\bigr),
\label{eq:uav_logits}
\end{equation}
Note that $\hat{\mathbf{S}}_u^{(t)}$ is an internal actor feature and does not belong to the formal observation $\boldsymbol{o}_u^{\mathrm{M}(t)}$. It feeds into the Pred‑CHGA alignment score $\varphi_p^{(t)}$ (Eq.~\eqref{eq:alpha_pred}).

\textbf{3) Centralized critic.}
\label{sec:method:critic}
\label{sec:method:deploy}
The critic \(V_\psi(s_{\mathrm{c}}^{(t)})\), with parameters \(\psi\) as in Sec.~\ref{sec:related:prelim}, is employed exclusively during training (Fig.~\ref{fig:system_arch}, block~S2) to compute the generalized‑advantage estimates required for the gradient estimator in Eq.~\eqref{eq:team_policy_grad}.
Its input concatenates the network‑wide summary from Eq.~\eqref{eq:global_summary} with the fleet‑mean navigation vector:
\begin{equation}
s_{\mathrm{c}}^{(t)}
=
\bigl({g_u^{\mathrm{G}}}^{(t)};\,
\operatorname{mean}_{u}\mathbf{z}_u^{(t)}\bigr).
\label{eq:critic_summary}
\end{equation}
The fleet‑mean navigation term remains critic‑only. Each deployed UAV can observe its own local $\mathbf{z}_u^{(t)}$, yet it cannot access fleet‑aggregated statistics.
This additional aggregate feature enables the critic to evaluate fleet‑wide relay‑fleet congestion and mitigates variance in GAE estimates.

\subsection{PPO Training}
\label{sec:method:training}
\label{sec:method:objective}
Based on the decentralized actors and centralized critic described above, we adopt PPO~\cite{schulman2017proximal} as our on‑policy optimizer.
Training utilizes the extended reward \(r_{\mathrm{ext}}^{(t)}\) defined in Sec.~\ref{sec:method:reward}, which reduces to \(r^{(t)}\) when CHGA is disabled.

This subsection details the Rollout Buffer block~(S3) and the PPO Update block~(S4) in Fig.~\ref{fig:system_arch}.
Filling the buffer still uses the observations and actors--critic of blocks~(S1)--(S2): on‑policy trajectories are stored in the buffer, after which GAE and the PPO update operate on the stored tuples.

Given buffered transition tuples \((\mathbf{o}^{(t)},\mathbf{a}^{(t)},r_{\mathrm{ext}}^{(t)},s_{\mathrm{c}}^{(t)})\), the centralized critic \(V_\psi(s_{\mathrm{c}}^{(t)})\) computes GAE advantages \(\hat{A}^{(t)}\) from the stored extended‑reward trajectory as in Sec.~\ref{sec:related:prelim}.
We then evaluate the clipped PPO surrogate \(L_{\mathrm{clip}}(\theta)\) from Eq.~\eqref{eq:prelim_ppo_clip} using the probability ratio given in Eq.~\eqref{eq:prelim_ppo_ratio}.
The critic target corresponds to the fixed return
\begin{equation}
\hat{V}_{\mathrm{ret}}^{(t)}=\hat{A}^{(t)}+V_t,
\label{eq:gae_return}
\end{equation}
where \(V_t\) denotes the critic value stored alongside each transition tuple. We define the value‑function loss and policy entropy loss as
\begin{align}
L_V &= \mathbb{E}\!\left[\bigl(V_\psi(s_{\mathrm{c}}^{(t)})-\hat{V}_{\mathrm{ret}}^{(t)}\bigr)^2\right], \label{eq:critic_loss}\\
L_H &= \mathbb{E}\!\left[H\!\left(\pi_\theta(\cdot\mid\mathbf{o}^{(t)})\right)\right],
\label{eq:entropy_loss}
\end{align}
respectively. Given positive weighting coefficients \(c_v>0\) and \(c_e>0\), the base PPO loss is formulated as
\begin{equation}
L_P = -L_{\mathrm{clip}} + c_v L_V - c_e L_H.
\label{eq:ppo_base}
\end{equation}

Let \(\mathcal{T}_h\subset\mathbb{N}\) be the set of prediction horizons and \(w_\Delta\ge 0\) the weight assigned to each horizon \(\Delta\in\mathcal{T}_h\). For Pred‑CHGA, we introduce a multi‑horizon auxiliary loss \(L_p\) to regularize sector‑congestion forecasts. This loss regresses the predicted sector‑congestion vector \(\hat{\mathbf{S}}_u^{(t)}\) against the future ground‑truth sector‑congestion \(\mathbf{S}_u^{(t+\Delta)}\) for every horizon \(\Delta\in\mathcal{T}_h\):
\begin{align}
L_p^{(\Delta)} &=
\frac{1}{N_{\mathrm{uav}}}\sum_{u=0}^{N_{\mathrm{uav}}-1}
\left\|
\hat{\mathbf{S}}_u^{(t)}-\mathbf{S}_u^{(t+\Delta)}
\right\|_2^2, \nonumber\\
L_p &= \sum_{\Delta\in\mathcal{T}_h} w_\Delta L_p^{(\Delta)}.
\label{eq:aux_loss}
\end{align}

Combining the base PPO loss with this weighted auxiliary term yields the overall loss to be minimized in each PPO update step:
\begin{equation}
L = L_P + \lambda L_p.
\label{eq:total_loss}
\end{equation}
Here, \(\lambda\ge 0\) controls the contribution of \(L_p\). Setting \(\lambda=0\) recovers the vanilla base loss \(L=L_P\). At each training epoch, the network parameters \((\theta,\psi)\) are updated using learning rate \(\eta\) according to this composite loss.

\section{Simulation Results and Discussions}
\label{sec:experiments}
\raggedbottom
\subsection{Experimental Setup and Evaluation Metrics}
\label{sec:exp:settings}
\makeatletter
\global\@nobreakfalse
\clubpenalty=200
\makeatother
\paragraph{Experimental Setup.}
\Judor{} is implemented in Python, with its environment interface built upon Gymnasium~\cite{brockman2016openai}. Model training is performed via the Tianshou~\cite{weng2022tianshou} library using vectorized parallel environments that instantiate the system described in Secs.~\ref{sec:system}--\ref{sec:method}.
We define a \textit{\Judor{} variant} as a concrete instantiation of the core framework with a specific configuration for ablation‑study purposes.
Table~\ref{tab:exp_hparams} summarizes the baseline training and simulation settings used by \Judor{} variants.
\begingroup
\setlength{\intextsep}{3pt}
\begin{table}[t]
\caption{Baseline training and simulation settings for \Judor{} variants.}
\label{tab:exp_hparams}
\centering
\footnotesize
\setlength{\tabcolsep}{2.5pt}
\renewcommand{\arraystretch}{0.95}
\begin{tabular}{@{}l r @{\hspace{2pt}} r @{\hspace{0.7em}} l r @{\hspace{2pt}} r @{}}
\hline
Simulation scene & Value & Eq. & & Value & Eq. \\
\hline
Ground range $r_{\mathrm{gr}}$ & $300$\,m & \eqref{eq:comm_range} &
UAV range $r_{\mathrm{uav}}$ & $900$\,m & \eqref{eq:comm_range} \\
Horizon $T_{\max}$ & $5000$ & \eqref{eq:p1_delivery} &
TTL ref.\ $T_0$ & $4000$ & \eqref{eq:ttl_urgency} \\
Buffer $B_b$ & $10000$ & \eqref{eq:buf_util} &
Candidates $K$ & $128$ & \eqref{eq:routing_action} \\
Heading bins $B$ & $8$ & \eqref{eq:heading_action} &
Ground speed $v_{\mathrm{gr}}$ & $20$\,m/step & \\
UAV speed $v_{\mathrm{uav}}$ & $100$\,m/step & \eqref{eq:uav_motion} &
Top-$K_{\sigma}$ & $3$ & \eqref{eq:hotspot_block_dim} \\
Exchange decay $\omega_{\mathrm{ex}}$ & $0.02$ & \eqref{eq:neighbor_buf_mean} &
Payload bound $S_{\max}$ & $4$ & \eqref{eq:candidate_feat} \\
Scene width $W$ & $10000$\,m & \eqref{eq:local_state} &
Scene height $H$ & $10000$\,m & \eqref{eq:local_state} \\
Default $N_{\mathrm{gr}}$ & $65$ & &
Default $N_{\mathrm{uav}}$ & $5$ & \\
\hline
PPO training & Value & Eq. &
Team reward \eqref{eq:team_reward} & Value & Eq. \\
\hline
Training epochs & $80$ & &
$\alpha_d$ & $4.0$ & \eqref{eq:p1_delivery} \\
Learning rate $\eta$ & $10^{-4}$ & &
$\alpha_e$ & $-1.5$ & \eqref{eq:p1_delivery} \\
Minibatch $B_{\mathrm{mb}}$ & $128$ & &
$\alpha_{dr}$ & $-0.5$ & \eqref{eq:p1_delivery} \\
Clip radius $\epsilon$ & $0.2$ & \eqref{eq:prelim_ppo_clip} &
$\alpha_b$ & $-0.1$ & \eqref{eq:p1_delivery} \\
Value loss $c_v$ & $0.5$ & \eqref{eq:ppo_base} &
$\alpha_{\rho}$ & $0.05$ & \eqref{eq:p1_uav} \\
Entropy $c_e$ & $0.01$ & \eqref{eq:ppo_base} &
$\alpha_c$ & $0.015$ & \eqref{eq:p1_uav} \\
Train.\ envs & $8$ & &
$\alpha_x$ & $0.006$ & \eqref{eq:p1_uav} \\
Eval.\ envs $N_{\mathrm{eval}}$ & $10$ & \eqref{eq:d_test} &
$\alpha_k$ & $0.003$ & \eqref{eq:p1_uav} \\
Actor MLP width & $256$ & &
& & \\
\hline
\end{tabular}
\end{table}
\endgroup

These parameters fall into three groups and stay fixed across variants, except for the $N_{\mathrm{uav}}$ sweep in Table~\ref{tab:uav_num_sweep}:
\textbf{(i) PPO training hyper‑parameters:} training epochs, learning rate $\eta$, minibatch size $B_{\mathrm{mb}}$, PPO clip radius $\epsilon$ (Sec.~\ref{sec:related:prelim}), loss weights $c_v$, $c_e$, parallel environment counts, and actor MLP hidden‑layer width;
\textbf{(ii) simulation‑scene parameters:} the Helsinki‑medium WKT road network with default node count $N = N_{\mathrm{gr}} + N_{\mathrm{uav}} = 70$, TTL reference $T_0$, buffer capacity $B_b$, maximum forwarding candidate count $K$, and heading bin number $B$ as in Table~\ref{tab:exp_hparams} (only the fleet‑size sweep in Table~\ref{tab:uav_num_sweep} varies $N_{\mathrm{uav}}$);
\textbf{(iii) team‑reward weights:} coefficients of the team objective in Eq.~\eqref{eq:team_reward}, with gating coefficients $g_s$ and $g_m$ from Eq.~\eqref{eq:gates} distinct from the UAV context summary $g_u^{(t)}$, and $|\alpha_d|\gg|\alpha_{\rho}|,|\alpha_x|,|\alpha_k|$ so that $J(\pi)$ in Eq.~\eqref{eq:p1} remains delivery‑centric.
A comparatively large $B_b$ keeps buffer‑overflow drops rare relative to TTL expiry, so delivery is gated primarily by intermittent contacts and finite TTL.

Episodes are generated on‑policy within parallel training environments. All policy training and evaluation are conducted on an NVIDIA GeForce RTX 4090 GPU.

\paragraph{Traffic Modes.}
Four representative traffic modes M1--M4 are defined in Table~\ref{tab:traffic_modes}.
They span two orthogonal axes, source multiplicity and temporal injection, which respectively describe how many sources inject traffic and when those injections occur.
For each mode, the table reports the source-and-injection category, the generation rule, and the per-episode message count $M_{\mathrm{created}}$.
It also gives a scenario analogy that associates each mode with the message pattern of a realistic scenario, such as post-disaster rescue, sparse rural road networking, or remote sensing.
All modes share the Helsinki-medium road network, with destinations uniformly sampled over ground nodes. Owing to large differences in injected volume, absolute deliveries are compared only within the same mode.
\begingroup
\setlength{\intextsep}{3pt}
\begin{table}[t]
\caption{Traffic modes M1--M4 in the experimental evaluation.}
\label{tab:traffic_modes}
\centering
\small
\setlength{\tabcolsep}{1.8pt}
\renewcommand{\arraystretch}{1.1}
\resizebox{\columnwidth}{!}{%
\sloppy
\begin{tabular}{@{}>{\centering\arraybackslash}p{0.10\columnwidth}
  >{\raggedright\hspace{0pt}\arraybackslash}p{0.16\columnwidth}
  >{\raggedright\hspace{0pt}\arraybackslash}p{0.34\columnwidth}
  >{\raggedright\hspace{0pt}\arraybackslash}p{0.34\columnwidth}
  c@{}}
\hline
Mode & \makecell[l]{Source /\\ injection} & \makecell[l]{Injection\\ rule} & \makecell[l]{Simulation\\ scenario} & $M_{\mathrm{created}}$ \\
\hline
M1 & Single-source; one-shot &
At $t{=}0$, one source injects 69 messages to random ground destinations. &
A rescue center sends maps and task orders to many field teams at once. &
69 \\
\hline
M2 & Full multi-source;\newline sustained &
Every 20 steps, all ground nodes inject messages with random destinations. &
In a farm IoT monitoring scenario or a long rescue scenario, every node keeps sending sensor readings. &
$\sim$4321 \\
\hline
M3 & Partial multi-source;\newline sustained &
Every 20 steps, a random ground-node subset injects asynchronously. &
Several rescue teams send status updates from different places over time. &
$\sim$306 \\
\hline
M4 & Multi-source; one-shot &
At $t{=}0$, five sources inject concurrently with random destinations. &
Several hard-hit places send emergency calls at the same time. &
$\sim$345 \\
\hline
\end{tabular}%
}
\end{table}
\endgroup

\paragraph{Evaluation Metrics.}
After each training epoch $e$, held‑out evaluation yields the mean episodic delivery across $N_{\mathrm{eval}}$ evaluation environments:
\begin{equation}
D_{\text{test}}^{(e)}
=
\frac{1}{N_{\mathrm{eval}}}\sum_{n=0}^{N_{\mathrm{eval}}-1} D_n^{(e)},
\label{eq:d_test}
\end{equation}
where $N_{\mathrm{eval}}$ denotes the number of parallel evaluation environments given in Table~\ref{tab:exp_hparams}, and $D_n^{(e)}$ is the number of successfully delivered messages over one full $T_{\max}$‑length episode in evaluation environment $n$.
Higher $D_{\text{test}}^{(e)}$ corresponds to improved delivery performance.

For \Judor{} ablation studies and UAV‑fleet‑size sweeps (Tables~\ref{tab:ablation_r1_merged}, \ref{tab:uav_num_sweep}, Fig.~\ref{fig:four_traffic_ablation}), we report peak and terminal held‑out delivery values:
\begin{equation}
D_{\text{peak}}
=
\max_{e\in\{1,\ldots,80\}} D_{\text{test}}^{(e)},
\qquad
D_{\text{term}}
=
D_{\text{test}}^{(80)},
\label{eq:d_peak_term}
\end{equation}
where larger values of both metrics are desirable.

For comparisons against routing‑only baselines (Table~\ref{tab:baselines}), we report delivery and expiry $(D,E)$ from a single rollout, alongside the delivery ratio
\begin{equation}
\eta_{\mathrm{del}}
=
\frac{D}{M_{\mathrm{created}}},
\label{eq:eta_del}
\end{equation}
in which $M_{\mathrm{created}}$ denotes the per‑mode total message count specified in Table~\ref{tab:traffic_modes}.
Higher $D$ and $\eta_{\mathrm{del}}$, alongside lower $E$, indicate better performance.
\Judor{} rows report $D_{\text{peak}}$ after training.

All subsequent subsections analyze experimental results using the setup and evaluation metrics defined above.
\FloatBarrier

\subsection{Ablation Study}
\label{sec:exp:m1_ablation}
\label{sec:exp:high_load}

Ablation experiments evaluate \(D_{\text{peak}}\) and \(D_{\text{term}}\) in Eq.~\eqref{eq:d_peak_term}.
Table~\ref{tab:exp_stacks} defines every named variant and lists only the configuration that differs from \emph{base}.
Table~\ref{tab:ablation_r1_merged} reports the cross-traffic results under M1--M4.
The comparison covers observation factorization, CHGA auxiliary supervision, and joint UAV--routing control.
\begin{table}[!t]
\caption{Named variants in the ablation study.
\emph{base} is the reference. Only the configuration that differs from it is listed.}
\label{tab:exp_stacks}
\centering
\footnotesize
\setlength{\tabcolsep}{3pt}
\renewcommand{\arraystretch}{1.12}
\begin{tabular}{@{}>{\raggedright\arraybackslash}p{0.32\linewidth}>{\raggedright\arraybackslash}p{0.64\linewidth}@{}}
\hline
Variant & Configuration \\
\hline
\emph{base} & ${g_u^{\mathrm{G}}}^{(t)}$, full $\mathbf{z}_u^{(t)}$, $\lambda{=}0$, alignment off \\
base local-ctx & ${g_u^{\mathrm{L}}}^{(t)}$; forecast branch at $\lambda{=}0.05$ \\
deploy-near-real & ${g_u^{\mathrm{L}}}^{(t)}$; $\mathbf{z}_u^{(t)}$ retains only $\mathbf{S}_u^{(t)}$; $\lambda{=}0$ \\
base $-$ UAV & $N_{\mathrm{uav}}{=}0$ \\
base + GT-CHGA & $(\alpha_a,\alpha_{ad},\alpha_{ar})=(0.003,0.012,0.004)$; $\lambda{=}0$ \\
base + Pred-CHGA & $\lambda{=}0.05$, $\mathcal{T}_h{=}\{1,10,25\}$, $w_\Delta\in\{0.34,0.33,0.33\}$, $(\alpha_a,\alpha_{ad},\alpha_{ar})=(0.003,0.012,0.004)$ \\
\hline
\end{tabular}
\end{table}

\begin{figure*}[!t]
\centering
\IfFileExists{figures/fig_four_traffic_ablation.png}{%
  \includegraphics[width=\textwidth]{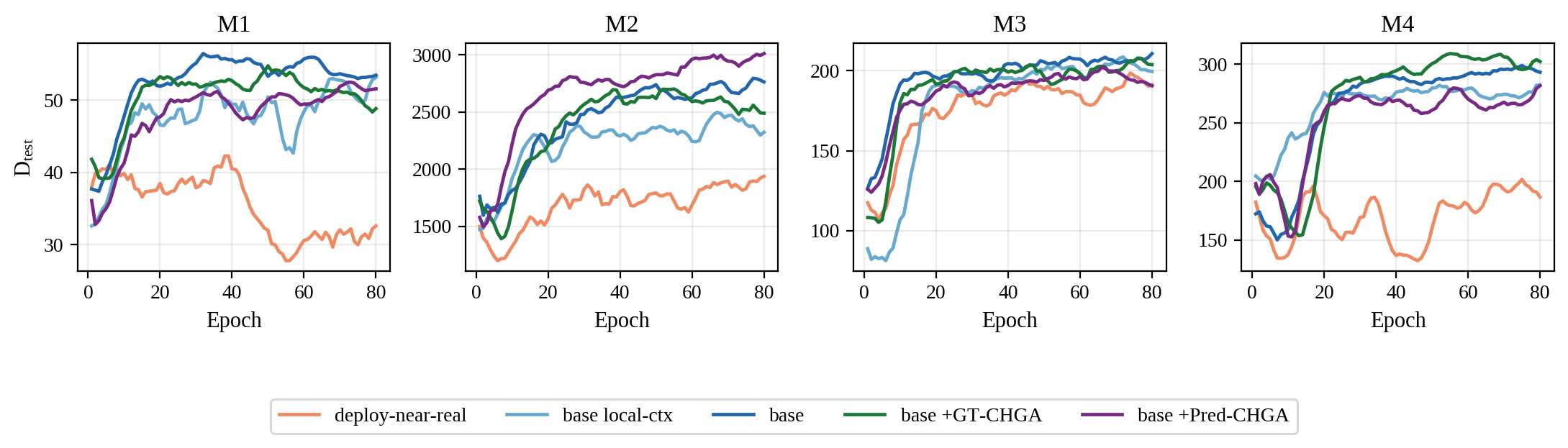}}{%
  \fbox{\parbox{0.9\linewidth}{\scriptsize Run \texttt{build\_four\_traffic\_ablation\_figure.py}}}}
\caption{Held-out $D_{\text{test}}$ versus training epoch.
Each panel uses an independent $y$-axis scale.}
\label{fig:four_traffic_ablation}
\label{fig:ablation_node1_curves}
\label{fig:m1_learning_curves}
\end{figure*}

Fig.~\ref{fig:four_traffic_ablation} shows the observation and CHGA learning curves.
The fleet-size sweep is reported separately in Table~\ref{tab:uav_num_sweep}.
Table~\ref{tab:ablation_r1_merged} reports $D_{\text{peak}}$ and $D_{\text{term}}$ for the same runs.
\begin{table}[!t]
\caption{Cross-traffic ablation on Helsinki-medium under M1--M4.
Each cell is $D_{\text{peak}}/D_{\text{term}}$ (Eq.~\eqref{eq:d_peak_term}).
The configuration of each variant is given in Table~\ref{tab:exp_stacks}.}
\label{tab:ablation_r1_merged}
\centering
\footnotesize
\setlength{\tabcolsep}{3.5pt}
\renewcommand{\arraystretch}{1.12}
\begin{tabular*}{\columnwidth}{@{\extracolsep{\fill}}l cccc@{}}
\hline
Configuration & M1 & M2 & M3 & M4 \\
\hline
\multicolumn{5}{@{}l}{\textit{A. Observation design}} \\
\emph{base} \emph{(ref.)} & \textbf{57.6/52.8} & 2948.8/2648.5 & 214.8/\textbf{214.8} & 302.8/\textbf{298.0} \\
base local-ctx & 54.8/52.0 & 2562.7/2366.8 & 211.8/198.6 & 290.8/282.2 \\
deploy-near-real & 49.3/31.0 & 2051.3/1989.0 & 203.1/195.0 & 228.0/184.8 \\
\hline
\multicolumn{5}{@{}l}{\textit{B. Congestion-hotspot-guided alignment}} \\
base + GT-CHGA & 55.4/51.5 & 2782.5/2438.8 & \textbf{217.0}/200.4 & \textbf{315.9}/289.4 \\
base + Pred-CHGA & 53.4/51.3 & \textbf{3120.8/3093.6} & 210.5/190.2 & 290.5/282.0 \\
\hline
\multicolumn{5}{@{}l}{\textit{C. Joint UAV control}} \\
base $-$ UAV & 13.2/11.8 & 815.5/788.2 & 70.9/68.5 & 98.2/90.8 \\
\hline
\end{tabular*}
\end{table}

\textbf{1) CTDE Information Factorization:}
\label{sec:exp:deploy_near_real}
Fig.~\ref{fig:four_traffic_ablation} and block~A of Table~\ref{tab:ablation_r1_merged} compare three actor observations, all trained with the centralized critic $s_{\mathrm{c}}^{(t)}$.
Alignment coefficients are $(\alpha_a,\alpha_{ad},\alpha_{ar})$.
$\lambda$, $\mathcal{T}_h$, and $w_\Delta$ follow Eqs.~\eqref{eq:total_loss} and~\eqref{eq:aux_loss}.
\emph{base} feeds the network-wide context ${g_u^{\mathrm{G}}}^{(t)}$.
\emph{base local-ctx} replaces it with the contact-limited context ${g_u^{\mathrm{L}}}^{(t)}$ of Eq.~\eqref{eq:local_summary} and keeps the forecast branch.
\emph{deploy-near-real} also uses ${g_u^{\mathrm{L}}}^{(t)}$, and additionally reduces $\mathbf{z}_u^{(t)}$ in Eq.~\eqref{eq:hotspot_block_dim} to the sector block $\mathbf{S}_u^{(t)}$ with the forecast branch off.
Contact-limited context with the navigation vector retained stays close to \emph{base} on every mode, but records a lower $D_{\text{peak}}$ and $D_{\text{term}}$ throughout, with the widest gap on M2.
Stripping the Top-$K_{\sigma}$ offset, the centroid direction, and the forecast widens the gap, most under bursty M4.
On M1, \emph{deploy-near-real} reaches a mid-training peak and then declines through the terminal epoch.
CTDE can therefore use $s_{\mathrm{c}}^{(t)}$ at training time while the deployed actors remain on contact-limited inputs, and the retained navigation features account for most of the remaining delivery.

\textbf{2) Congestion-Hotspot-Guided Alignment:}
\label{sec:exp:block_c}
Block~B of Table~\ref{tab:ablation_r1_merged} adds CHGA on top of \emph{base}, with the same alignment coefficients inside the extended reward $r_{\mathrm{ext}}^{(t)}$ of Eq.~\eqref{eq:reward_full}.
\emph{GT-CHGA} aligns headings with the ground-truth sector congestion $\mathbf{S}_u^{(t)}$, which enters only that reward.
\emph{Pred-CHGA} aligns headings with the forecast $\hat{\mathbf{S}}_u^{(t)}$, feeds the forecast into the heading MLP, and trains it at $\lambda{=}0.05$.
The CHGA gain varies with the traffic mode.
\emph{Pred-CHGA} exceeds both \emph{base} and \emph{GT-CHGA} on M2 in peak and terminal delivery.
\emph{GT-CHGA} raises only the M4 peak, and the terminal value stays below \emph{base} on most modes.

\textbf{3) Joint UAV Relaying and Fleet Scale:}
\label{sec:exp:uav_num_sweep}
Removing aerial relays (\emph{base $-$ UAV}, block~C of Table~\ref{tab:ablation_r1_merged}) drastically degrades delivery performance relative to all fleet‑equipped variants: for traffic mode M1, the peak delivery drops from 57.6 to 13.2.
Ground‑node communication contacts limited to $r_{\mathrm{gr}}{=}300$\,m cannot provide full coverage across the road network. By contrast, UAV links with $r_{\mathrm{uav}}{=}900$\,m dynamically reshape the contact matrix via Eq.~\eqref{eq:coupled_transition}.
Joint motion‑routing control thus constitutes a structural prerequisite for satisfactory performance under these scenarios.
Table~\ref{tab:uav_num_sweep} further investigates a separate configuration with swept $N_{\mathrm{uav}}$, where $\lambda{=}0.05$ and \emph{GT-CHGA} coefficients are held fixed.
\begin{table}[!t]
\caption{UAV fleet-size sensitivity on M1--M4.}
\label{tab:uav_num_sweep}
\centering
\scriptsize
\setlength{\tabcolsep}{2.5pt}
\renewcommand{\arraystretch}{1.05}
\begin{tabular*}{\columnwidth}{@{\extracolsep{\fill}}l cccc@{}}
\hline
$N_{\mathrm{uav}}$ & M1 & M2 & M3 & M4 \\
\hline
\multicolumn{5}{@{}l}{\textit{Cells: Peak/Term $D_{\text{test}}$}} \\
$N_{\mathrm{uav}}{=}0$ & 13.2/11.8 & 815.5/788.2 & 70.9/68.5 & 98.2/90.8 \\
$N_{\mathrm{uav}}{=}1$ & 45.7/45.7 & 1601.7/1397.0 & 133.0/119.0 & 106.6/96.9 \\
$N_{\mathrm{uav}}{=}2$ & 47.7/39.8 & 2014.7/1889.3 & 183.5/173.2 & 299.6/284.2 \\
$N_{\mathrm{uav}}{=}3$ & 49.8/44.7 & 2349.2/2349.2 & 181.0/162.0 & 275.2/238.5 \\
$N_{\mathrm{uav}}{=}4$ & 54.6/51.9 & 2860.4/2746.2 & 189.3/179.2 & 306.5/296.5 \\
$N_{\mathrm{uav}}{=}5$ (default) & 57.6/52.8 & 2948.8/2648.5 & 214.8/\textbf{214.8} & 302.8/298.0 \\
$N_{\mathrm{uav}}{=}6$ & 59.3/57.8 & 3032.9/\textbf{2991.1} & 215.2/201.1 & 302.4/257.8 \\
$N_{\mathrm{uav}}{=}7$ & 57.2/55.6 & \textbf{3047.3}/2853.1 & 210.8/201.9 & 304.1/\textbf{298.8} \\
$N_{\mathrm{uav}}{=}8$ & \textbf{60.9/58.2} & 2775.2/2667.3 & \textbf{224.1}/211.5 & \textbf{312.2}/291.5 \\
\hline
\end{tabular*}
\end{table}

Adding the first UAV yields the most substantial performance gain across all traffic modes.
Five UAVs already cover this map, so a larger fleet changes delivery only slightly.

Across traffic modes M1--M4, the primary performance improvements originate from the joint UAV‑routing co‑optimization enabled by the CTDE framework.
With $\mathbf{z}_u^{(t)}$ retained, \emph{base local-ctx} stays close to \emph{base} but remains slightly below it.
\emph{deploy-near-real} suffers a larger loss once this navigation feature vector is reduced to the sector block.
CHGA provides performance benefits only for certain traffic conditions.

\raggedbottom
\subsection{Baseline Routing Comparison}
\label{sec:comp:methods}
\label{sec:comp:results}
We consider five routing‑only baseline methods, each selecting at most one local transfer per node from the per‑step candidate sets defined in Sec.~\ref{sec:problem}:
\begin{enumerate}
  \item \textbf{PRoPHET:} Heuristic routing leveraging encounter history and transitive delivery predictability~\cite{lindgren2004probabilistic};
  \item \textbf{MaxProp:} Multi‑metric priority ranking based on destination recency, hop count, TTL urgency, and buffer occupancy~\cite{burgess2006maxprop};
  \item \textbf{Fan DPUVR:} Multi‑attribute multiplicative utility metric incorporating trajectory, energy, link persistence, distance, and queuing delay~\cite{fan2025novel};
  \item \textbf{ICC Q‑learning:} Tabular Q‑learning operating over $(\text{src},\text{relay},\text{dest})$ tuples with delivery‑oriented reward signals~\cite{dhurandher2020reinforcement};
  \item \textbf{ICC FQLRP:} The aforementioned Q‑learning backbone augmented with a fuzzy classification preprocessing layer for action selection~\cite{dhurandher2020reinforcement}.
\end{enumerate}

All five baselines are partially re‑implemented within our discrete‑time SCF simulation environment, sharing identical contact models and traffic‑mode configurations from Sec.~\ref{sec:exp:settings}, while preserving each algorithm’s core scoring or learning logic.
Performance is evaluated via single‑rollout metrics $(D,E)$ and delivery ratio $\eta_{\mathrm{del}}$ defined in Eq.~\eqref{eq:eta_del}.
In Table~\ref{tab:baselines}, \Judor{} cells are $D_{\text{peak}}$/$E$/$\eta_{\mathrm{del}}$, averaged over $N_{\mathrm{eval}}$ environments.
Among routing‑only approaches, PRoPHET achieves the overall best delivery performance across all traffic modes. Nevertheless, all heuristic and learning‑based baselines suffer degraded delivery ratios under the heavy sustained congestion imposed by M2.
MaxProp and Fan DPUVR yield only marginal performance improvements from their multi‑factor ranking rules. Meanwhile, the two tabular Q‑learning baselines are hampered by periodic Q‑table resets under continuous message injection.
By jointly optimizing UAV heading alongside forwarding decisions, \Judor{} outperforms the entire set of routing‑only baselines for every traffic scenario, including PRoPHET.

\begin{table}[H]
\centering
\caption{Routing-only baselines and selected \Judor{} peaks on Helsinki-medium under M1--M4 (Table~\ref{tab:traffic_modes}).
Routing-only cells are single-rollout $D$/$E$/$\eta_{\mathrm{del}}$ (Eq.~\eqref{eq:eta_del}); \Judor{} cells are $D_{\text{peak}}$/$E$/$\eta_{\mathrm{del}}$, averaged over $N_{\mathrm{eval}}$ environments, so $D$ and $E$ are fractional.
On M2 and M3, messages remain buffered at episode end, so $D+E<M_{\mathrm{created}}$.}
\label{tab:baselines}
\scriptsize
\setlength{\tabcolsep}{2.2pt}
\renewcommand{\arraystretch}{1.05}
\begin{tabular*}{\columnwidth}{@{\extracolsep{\fill}}l cccc@{}}
\hline
Method & M1 & M2 & M3 & M4 \\
\hline
\multicolumn{5}{@{}l}{\textit{A. Routing-only: $D$/$E$/$\eta_{\mathrm{del}}$}} \\
PRoPHET & 30.0/39.0/0.43 & 1611/371/0.37 & 157/17/0.51 & 207/138/0.60 \\
MaxProp & 10.0/59.0/0.14 & 911/553/0.21 & 92/30/0.30 & 104/241/0.30 \\
Fan DPUVR & 12.0/57.0/0.17 & 944/516/0.22 & 73/33/0.24 & 107/238/0.31 \\
ICC Q-learning & 17.0/52.0/0.25 & 706/593/0.16 & 65/43/0.21 & 85/260/0.25 \\
ICC FQLRP & 29.0/40.0/0.42 & 724/639/0.17 & 58/47/0.19 & 140/205/0.41 \\
\hline
\multicolumn{5}{@{}l}{\textit{B. \Judor{}: $D_{\text{peak}}$/$E$/$\eta_{\mathrm{del}}$}} \\
base & \textbf{57.6/11.3/0.83} & \textbf{2948.8/123.2/0.68} & \textbf{214.8/9.8/0.70} & \textbf{302.8/42.2/0.88} \\
base $-$ UAV & 13.2/55.8/0.19 & 815.5/642.7/0.19 & 70.9/43.4/0.23 & 98.2/246.8/0.28 \\
\hline
\end{tabular*}
\end{table}

\flushbottom
\section{Conclusion}
\label{sec:conclusion}
This paper has investigated joint UAV flight and opportunistic routing for delay‑tolerant networks under road‑constrained ground mobility, where system performance is hampered by intermittent contacts, finite node buffers, and sparse delayed delivery feedback.
We presented \Judor{}, a CTDE‑PPO framework equipped with factored routing and UAV heading actors for contact‑limited decentralized execution.
Our key contributions consist in jointly optimizing discrete UAV headings and per‑node SCF message replication, together with the optional CHGA module to alleviate weak delivery‑driven reward signals.
We extensively evaluated our framework under multiple representative traffic modes against both classical and learning‑based baseline protocols.
Simulation results demonstrate consistent delivery performance improvements over the competing baselines. Ablation studies reveal that most performance gains stem from controllable aerial relays and the CTDE observation design, whereas CHGA acts as a traffic‑dependent refinement instead of an indispensable component.

Despite these promising results, this work still has several limitations.
Our system model builds upon predefined ground mobility assumptions and a range-limited contact model. All results are obtained from discrete‑time simulations without real‑world flight validations.
Moreover, Pred‑CHGA cannot yield optimal performance across all traffic modes, and thus the auxiliary supervision mechanism should be tailored to target traffic modes.

For future work, we first intend to investigate moderate‑broadcast forwarding, allowing nodes to replicate one message to multiple neighbors per time slot.
Further, we will conduct outdoor testbed experiments for real‑world validation.
Additionally, we plan to adopt fine‑grained per‑message reward‑penalty shaping to mitigate sustained congestion, where heavier penalties are imposed on long‑residing messages and redundant copies, and evaluate our framework under diverse node mobility models.
The findings presented in this paper provide useful insights for follow‑up research on joint UAV flight and opportunistic routing in DTNs.

\bibliographystyle{IEEEtran}
\bibliography{refs}

\end{document}